\documentclass[lettersize,journal]{IEEEtran}
\usepackage{amsmath,amsfonts}
\usepackage{algorithmic}
\usepackage{algorithm}
\usepackage{array}
\usepackage[caption=false,font=normalsize,labelfont=sf,textfont=sf]{subfig}
\usepackage{textcomp}
\usepackage{stfloats}
\usepackage{url}
\usepackage{verbatim}
\usepackage{graphicx}
\usepackage{cite}
\usepackage{caption}
\usepackage[table,xcdraw]{xcolor} 
\usepackage{graphicx}
\usepackage{amsmath}
\usepackage{amssymb}
\usepackage{booktabs}
\usepackage{tikz}
\usetikzlibrary{mindmap} 
\usepackage{arydshln}
\usepackage[margin=1in]{geometry}
\usepackage[normalem]{ulem} 
\usepackage{rotating} 
\usepackage{tablefootnote}
\useunder{\uline}{\ul}{}
\usepackage{adjustbox}
\usepackage{bm}

\usepackage[accsupp]{axessibility} 

\definecolor{JINZAMOMI}{RGB}{225,122,119} 
\definecolor{AKEBONO}{RGB}{241,148,131} 
\definecolor{TOKI}{RGB}{238,169,169} 
\definecolor{MOGEI}{RGB}{123,162,63} 
\definecolor{HIWA}{RGB}{190,194,63} 
\definecolor{NOSHIMEHANA}{RGB}{43,95,117}
\definecolor{KUCHINASHI}{RGB}{246,197,85} 
\definecolor{USUKI}{RGB}{250,214,137} 
\definecolor{HANAASAGI}{RGB}{30,136,168} 
\definecolor{SORA}{RGB}{88,178,220} 
\definecolor{NAE}{RGB}{134,193,102} 
\definecolor{KOKE}{RGB}{131,138,45} 
\definecolor{WASURENAGUSA}{RGB}{125,185,222} 
\definecolor{GUNJYO}{RGB}{81,168,221} 
\definecolor{KAMENOZOKI}{RGB}{165,222,228} 

\usepackage[pagebackref,breaklinks,colorlinks]{hyperref}

\usepackage[capitalize]{cleveref}
\crefname{section}{Sec.}{Secs.}
\Crefname{section}{Section}{Sections}
\Crefname{table}{Table}{Tables}
\crefname{table}{Tab.}{Tabs.}

\usepackage{multirow}
\usepackage{fontawesome}
\usepackage{colortbl}

\usepackage{tcolorbox}
\usepackage{xpatch}
\usepackage{diagbox}
\usepackage{utfsym}

\usepackage{amsfonts}
\usepackage{mathrsfs}

\usepackage[T1]{fontenc} 
\pdfmapline{+delphine < Delphine.ttf <T1-WGL4.enc}

\usepackage{MnSymbol}
\usepackage{makecell}

\usepackage{placeins} 
\usepackage{float} 
\usepackage{graphicx} 
\usepackage{adjustbox} 

\begin{document}

\title{DeepAndes: A Self-Supervised Vision Foundation Model for Multi-Spectral Remote Sensing Imagery of the Andes}

\author{
Junlin~Guo,
James~R.~Zimmer-Dauphinee,
Jordan~M.~Nieusma,
Siqi~Lu,
Quan~Liu,\\
Ruining~Deng,
Can~Cui,
Jialin~Yue,
Yizhe~Lin,
Tianyuan~Yao,\\
Juming~Xiong,
Junchao~Zhu,
Chongyu~Qu,
Yuechen~Yang,
Mitchell~Wilkes,\\
Xiao~Wang,
Parker~VanValkenburgh,
Steven~A.~Wernke,
and Yuankai~Huo
\thanks{Junlin Guo, Siqi Lu, Jialin Yue, Juming Xiong, Chongyu Qu, Mitchell Wilkes, and Yuankai Huo are with the Department of Electrical and Computer Engineering, Vanderbilt University, Nashville, TN, USA. \textcolor{black}{(Corresponding author: Yuankai Huo.)}}%
\thanks{James R. Zimmer-Dauphinee and Steven A. Wernke are with the Department of Anthropology, Vanderbilt University, Nashville, TN, USA.}%
\thanks{Jordan M. Nieusma and Yuankai Huo are with the Data Science Institute, Vanderbilt University, Nashville, TN, USA.}%
\thanks{Quan Liu, Ruining Deng, Can Cui, Tianyuan Yao, Junchao Zhu, Yuechen Yang, and Yuankai Huo are with the Department of Computer Science, Vanderbilt University, Nashville, TN, USA.}%
\thanks{Yizhe Lin is with the Department of Mathematics, Vanderbilt University, Nashville, TN, USA.}%
\thanks{Xiao Wang is with Oak Ridge National Laboratory, Oak Ridge, TN, USA.}%
\thanks{Parker VanValkenburgh is with the Department of Anthropology, Brown University, Providence, RI, USA.}%
\thanks{Ruining Deng is also with the Department of Radiology, Weill Cornell Medicine, New York, NY, USA.}%
}

\maketitle

\begin{abstract}

By mapping sites at large scales using remotely sensed data, archaeologists can generate unique insights into long-term demographic trends,  inter-regional social networks, and human adaptations in the past. Remote sensing surveys complement field-based approaches, and their reach can be especially great when combined with deep learning and computer vision techniques. However, conventional supervised deep learning methods face challenges in annotating fine-grained archaeological features at scale. In addition, while recent vision foundation models have shown remarkable success in learning large-scale remote sensing data with minimal annotations, most off-the-shelf solutions are designed for RGB images rather than multi-spectral satellite imagery, such as the 8-band data used in our study. In this paper, we introduce DeepAndes, a transformer-based vision foundation model trained on three million multi-spectral satellite images, specifically tailored for Andean archaeology. DeepAndes incorporates a customized DINOv2 self-supervised learning algorithm optimized for 8-band multi-spectral imagery, marking the first foundation model designed explicitly for the Andes region. We evaluate its image understanding performance through imbalanced image classification, image instance retrieval, and pixel-level semantic segmentation tasks. Our experiments show that DeepAndes achieves superior F1 scores, mean average precision, and Dice scores in few-shot learning scenarios, significantly outperforming models trained from scratch or pre-trained on smaller datasets. This underscores the effectiveness of large-scale self-supervised pre-training in archaeological remote sensing. Codes will be available on https://github.com/geopacha/DeepAndes.
\end{abstract}

\begin{IEEEkeywords}
Foundation Model, Self-supervised Learning, DINOv2, Multi-Spectral Imaging, Remote Sensing, Andean Archaeology
\end{IEEEkeywords}

\section{Introduction}
\label{sec:introduction}

\IEEEPARstart{O}{ne} of the most vexing and persistent challenges for field-based sciences such as archaeology, population biology, demography, environmental monitoring, and field geology is conducting analyses at large scales. At the level of small regions, conventional field-based survey methods have proven to be highly effective. \textcolor{black}{Through such surveys, field scientists reconstruct and analyze populations, environments, and resources at regional levels and provide crucial data for modeling them at larger scales \cite{proposal_8, proposal_9, proposal_10, proposal_11, proposal_12, tripcevich2010site, fuller1998integration, allwine2002overview}. However, generating accurate datasets that record continuous distributions (particularly of highly variable cultural phenomena, such as archaeological sites and the distribution of modern urban areas) has proven difficult to achieve at inter-regional and continental scales without large resource outlays to fund multi-year field campaigns. \cite{proposal_19, proposal_20, proposal_21, vanvalkenburgh2020big}.} These problems are especially notable in Andean South America, where the the topography and inaccessibility of many areas make field surveys challenging. Since the early 2000s, field scientists have used high-resolution remote sensing satellite imagery to analyze the distribution of archaeological features, natural resources, and modern populations at larger scales. However, such ``brute force'' manual imagery surveys remain labor intensive, time consuming, and prone to observational fatigue and inter-observer variability in feature detection \cite{casana2014regional}. Developing effective AI-assisted approaches for autonomous information extraction will enable us to expand survey coverage at scale, providing new insights into human adaptation, settlement patterns, landscapes, and networks in the Andes.

With the rapid advancement of artificial intelligence, deep learning has made significant contributions in a diverse series of domains \cite{ke2024tshfna, li2024u, yue2025glofinder, sharma2021machine, ssl_prior, zhu2024asign}. \textcolor{black}{Recently, the emergence of \textbf{foundation models} (FMs) has further expanded the scope of deep learning applications \cite{zhou2024comprehensive, cui2024pfps, guo2025assessment, guo2024good, awais2025foundation, cui2024enhancing}, including remote sensing~\cite{liu2024remoteclip,lu2025vision, li2025unleashing}. As summarized in \cite{lu2025vision}, foundation models pretrained on massive remote sensing datasets demonstrate strong adaptability across a wide range of downstream tasks in both earth and social sciences. As a domain-specific application, archaeological remote sensing has particularly benefited from foundation models, which have shown promise in tasks such as artifact recognition \cite{ayush2021geography}, detection of archaeological structures \cite{feng2023self}, and texture analysis \cite{akiva2022self}, offering new opportunities for large-scale archaeological analysis. Despite these advances, most of the earth's surface has yet to be systematically surveyed, and vast regions are underrepresented in archaeological research, including the Andes region. High levels of inter-regional variability in land cover and the diversity of archaeological features themselves have proven to be major barriers to achieving such coverage. Thus, developing a new AI foundation model with broad utility across the social and earth sciences in the Andes would be highly valuable, enabling experts to contribute where they are most effective—as observers and analysts in the archaeological workflow.}

\begin{figure*} [!htp]
\begin{center}
\includegraphics[width=1\linewidth]{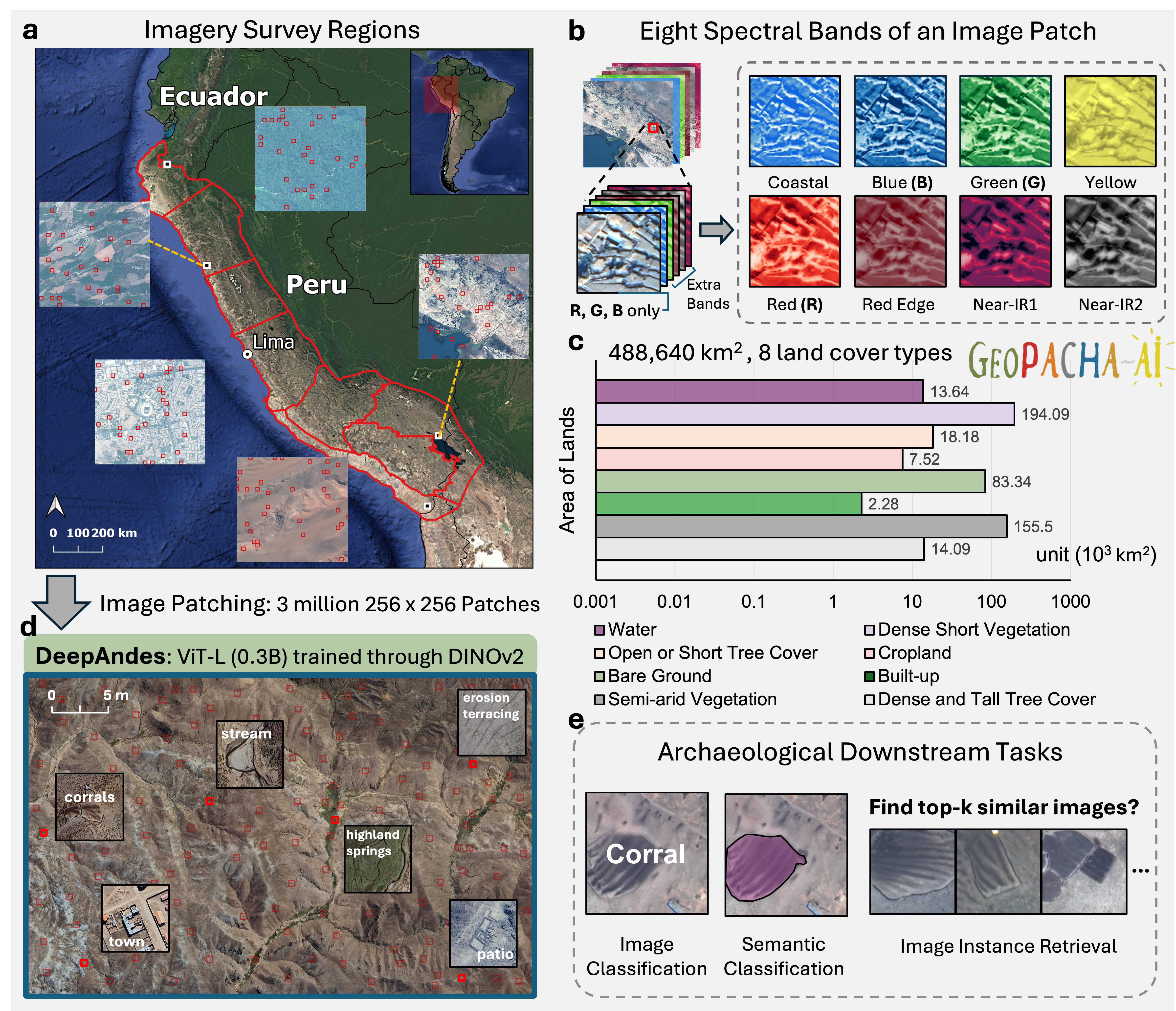}
\end{center}
\caption{\textbf{Overview of DeepAndes}. This figure shows the training dataset (\textbf{a-d}) and three domain-specific downstream tasks (\textbf{e}) using DeepAndes — a vision foundation model designed for multi-spectral satellite imagery in the Andes region. Particularly, \textbf{(a)} shows a large-scale map of the imagery used to train DeepAndes, highlighting various land cover types, with their area distribution shown in \textbf{(c)}. \textbf{(b)} presents the unit sample patch (\textcolor{red}{red box} in a, b, d) with eight spectral bands. \textbf{(d)} illustrates image patching for DINOv2 training, with geospatial sampling densely covering different archaeological sites.}
{ \label{fig:fig1}}
\end{figure*}

To bridge this gap, this paper proposes DeepAndes, the first vision foundation model for remote sensing of the Andean region. \textbf{\textit{To the best of our knowledge, no foundation model has been specifically developed for the Andes using 8-channel multi-spectral satellite imagery.}} As shown in Fig.~\ref{fig:fig1}, the multi-spectral training dataset, sourced from the central Andes (Figs.~\ref{fig:fig1}a-c), covers 488,640~km$^2$ and encompasses 8 distinct land cover types (Fig.~\ref{fig:fig1}c). DeepAndes, a 307M-parameter vision transformer (ViT) model, is trained on 3 million satellite image patches. A key challenge in developing a novel vision foundation model learning framework for high-dimensional multi-spectral remote sensing images is that most off-the-shelf vision foundation models are primarily designed for three-band (RGB) imagery. To address this problem, we customized the DINOv2 \cite{oquab2023dinov2} self-supervised learning (SSL) pipelines to support our 8-band multi-spectral imagery. During pre-training, we also adjusted the scale of global-local image views to align with the size of architectural and archaeological features in the dataset. As shown in Fig.~\ref{fig:fig1}e, to evaluate DeepAndes, we investigated its image pattern recognition and few-shot learning capabilities through three key vision-based remote sensing tasks: imbalanced image classification, image instance retrieval, and image segmentation. These downstream tasks are crucial for large-scale remote sensing datasets, where data are often highly heterogeneous and features are sparsely distributed. Similarly to \cite{sakai2024ai_pnas}, DeepAndes-based image instance retrieval can help identify patterns in the distribution of cultural and natural features within vast datasets, improving the efficiency of large-scale archaeological analysis.

\begin{figure*} [ht]
\begin{center}
\includegraphics[width=0.7\linewidth]{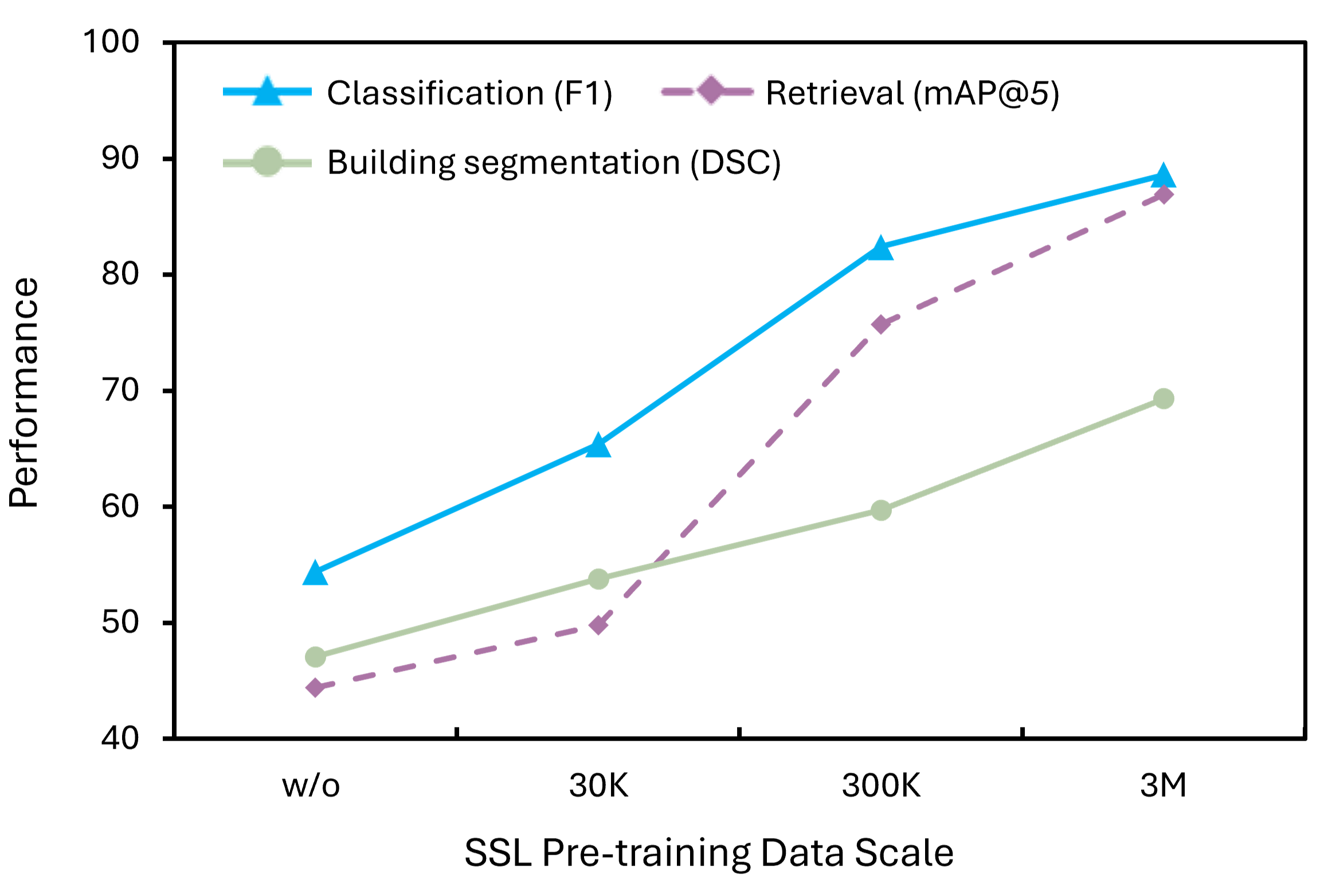}
\end{center}
\caption{\textbf{Scaling law is observed in DeepAndes.} This figure illustrates the model’s performance across three key downstream tasks: site classification, image retrieval, and building segmentation. The results are presented for models trained with no pretraining (w/o), 30K, 300K, and 3 million images. The findings highlight the scalability of DeepAndes, indicating that its performance can be further improved with larger training datasets.}
{ \label{fig:scale_law}}
\end{figure*} 

Through extensive experiments employing archaeological features as a test case, our results demonstrate that self-supervised pre-training enhances DeepAndes's performance in both image-level and pixel-level downstream tasks. Among the imbalanced archaeological loci (discrete archaeological features of interest) classification task, DeepAndes achieves an F1 score of 0.886. Even when fine-tuned on only 10\% of the original classification data, it maintains a strong F1 score of 0.83, substantially outperforming models trained from scratch using supervised methods. For the archaeological image instance retrieval task, features extracted from the frozen DeepAndes backbone achieve a mean average precision (mAP) of 0.869 within the top-\textit{k} retrieved samples (mAP@$k$, k=5), demonstrating the effectiveness of the feature representations. For pixel-level dense feature recognition, we evaluated segmentation tasks across three challenging archaeological loci, where the diversity of locus (singular form of loci) objects adds complexity.  Our few-shot learning results show that DeepAndes excels in both transfer learning and fine-tuning settings when using a simple linear segmentation head. Interestingly, the scaling law is observed from the DeepAndes model as depicted in Fig.~\ref{fig:scale_law}. In short, DeepAndes, pre-trained on 3 million diverse remote sensing satellite image patches, outperforms the same ViT models pre-trained on smaller-scale datasets (e.g., 30K and 300K images) across all downstream tasks, underscoring the effectiveness of large-scale self-supervised learning in this work.

The contributions of this paper are three-fold:
\begin{itemize}
  \item We propose DeepAndes, the first AI-driven vision foundation model for the Andean region, leveraging large-scale pre-training on multi-spectral high-resolution satellite images for downstream analysis in field sciences.

  \item We revise the DINOv2 framework for multi-spectral pre-training using 8-band WorldView-2 (WV-2) and Worldview-3 (WV-3) satellite imagery. This modified framework leverages the flexibility of DINOv2 with data preprocessing, data augmentation, and network architecture, enabling seamless adaptation to other multi-channel remote sensing data in the future, regardless of the number of input channels.
  
  \item We demonstrate the few-shot adaptability of DeepAndes to downstream applications, showcasing its effectiveness across three prevalent archaeological remote sensing tasks: imbalanced image classification, image instance retrieval, and pixel-level image segmentation. 
  
\end{itemize}

\section{Related work}
In this section, we describe the background of this study, covering foundation models in remote sensing and summarizing two major SSL strategies for foundation model pre-training.

\subsection{Foundation Models For Remote Sensing}
The emergence of foundation models has revolutionized the field of remote sensing through their capacity to serve as versatile pre-trained frameworks for various downstream applications \cite{lu2025vision}. These models excel in processing complex remote sensing data, including multi-spectral and multi-temporal imagery, by leveraging extensive pre-training on large-scale datasets \cite{Jiao2023}. The integration of SSL approaches \cite{jing2019selfsupervised} and transformer architectures \cite{vaswani2023attention} has substantially improved performance across tasks such as image classification and change detection \cite{Dias2023}.
A distinctive advantage of FMs in remote sensing lies in their capability to extract meaningful representations from unlabeled data through SSL techniques \cite{zhou2024comprehensive}. The incorporation of transformer architectures \cite{vaswani2023attention} enables these models to effectively process geospatial data's unique characteristics, including variable spatial resolutions and temporal patterns, building upon earlier findings \cite{lu2025vision}.
The development trajectory of FMs has been shaped by both deep learning advances and data availability. While early progress centered on CNN architectures like ResNet \cite{he2015deep} for image analysis tasks \cite{Ma2024}, the advent of transformer models has significantly enhanced the processing of large-scale imagery \cite{SatMAE}. ViT models have particularly advanced the field by processing images as sequences of patches, enabling comprehensive analysis of both local details and global patterns. Recent advancements in FMs are transforming remote sensing by improving representation learning for diverse geospatial tasks. Notable examples include SatMAE \cite{SatMAE}, designed for temporal and multi-spectral satellite imagery analysis; Scale-MAE \cite{Scale-MAE}, which focuses on multiscale geospatial representation learning; SkySense \cite{guo2024skysense}, a billion-scale pre-trained universal model for earth observation imagery; and DINO-MC \cite{DINO-MC}, which enhances SSL capabilities for remote sensing applications. However, these advances face ongoing challenges, including data quality requirements, computational demands, and domain adaptation needs \cite{SatMAE++}.

\subsection{Self-Supervised Representation Learning Strategies}
In the field of remote sensing and computer vision, self-supervised learning represents a fundamental component in foundation model pre-training \cite{lu2025vision}. During the pre-training stage, SSL allows models to learn meaningful representations without relying on labeled datasets. This capability is particularly beneficial in remote sensing, where labeled data is often scarce. Models pre-trained using SSL are adept at identifying patterns and features in large volumes of unlabeled remote sensing data, making them highly effective for various downstream applications. As summarized in \cite{lu2025vision}, commonly used SSL techniques in remote sensing FMs can fall into two main categories: contrastive learning and predictive coding.

\textbf{Contrastive Learning.} Contrastive learning focuses on learning data representations by comparing different augmented versions of the same data point. It pulls similar (positive) pairs closer and push dissimilar (negative) pairs apart in the representation space. The methodology depends on advanced data augmentation techniques, such as random cropping, rotation, and color jittering, to create diverse views of the same image. Notable implementations that have shown success in remote sensing applications include DINO \cite{DINO}, which employs self-distillation, SimCLR \cite{SimCLR}, which uses a simple contrastive framework, and MOCO (Momentum Contrast) \cite{MoCo}, which leverages a dynamic dictionary approach \cite{lu2025vision}.

\textbf{Predictive Coding.} Predictive coding approaches, as discussed in \cite{lu2025vision}, focus on training models to reconstruct missing input data components from available observations. This methodology has proven particularly effective in capturing both spatial and temporal relationships within remote sensing data, especially when dealing with multi-spectral imagery and cloud-covered regions \cite{lu2025vision}. The strategy commonly employs architectures such as autoencoders (AE), which learn compressed representations through encoding-decoding processes, and masked autoencoders (MAE), which specifically focus on reconstructing masked portions of input data \cite{lu2025vision}. These approaches have demonstrated significant success in developing robust internal representations of complex remote sensing data structures.

This work employs one of the SOTA SSL pre-training strategies, DINOv2~\cite{oquab2023dinov2}, on million-scale satellite image datasets from the Central Andes. The DINOv2 algorithm leverages contrastive learning concepts with knowledge distillation. Additionally, the ``masking'' concept is utilized with iBOT \cite{zhou2021ibot} loss to further encourage patch-level representation learning.


\section{Methodologies}
\label{sec:methodologies}
This section details the methodologies employed in this work. It begins with the construction of the million-scale pre-training dataset, followed by an overview of the DINOv2 framework and our pre-training strategies. Finally, it presents the downstream analysis using the pre-trained DeepAndes backbone.

\subsection{Million-Scale Training Dataset Construction}
\label{methods:train_dataset}

As described previously in Figure~\ref{fig:fig1}, the dataset used for SSL pre-training was sourced from across the Peruvian Andes, surveyed from 6 teams, spanning 488,640 $km^{2}$, and including 8 distinct land cover types. Our high-resolution satellite imagery is from the WorldView-2 and WorldView-3 satellites, both of which are composed of eight multi-spectral bands, plus a panchromatic band. After pansharpening (fusing the panchromatic band with other spectral bands), WV-2 imagery is of a Ground Sample Distance (GSD) of 0.46 m (at nadir), while WV-3 imagery provides 0.31 m (at nadir) GSD resolution. Our imagery processing pipeline includes steps for image orthorectification, cloud removal, and GSD commensuration. Image processing includes upsampling WV-2 imagery to match the GSD of WV-3 imagery. To construct a diverse, large-scale pre-training dataset, we randomly sampled the Central Andean area (encompassing western Peru and parts of northern Chile and northwestern Bolivia) into non-overlapping 256 $\times$ 256 image patches (shown as the red box in Fig.~\ref{fig:fig1}). Featureless images with entirely dark or bright pixels, such as those containing only water or clouds, were removed. This process yielded a total of 3 million 256 $\times$ 256 image patches for pre-training DeepAndes.

\begin{figure*} [ht]
\begin{center}
\includegraphics[width=1\linewidth]{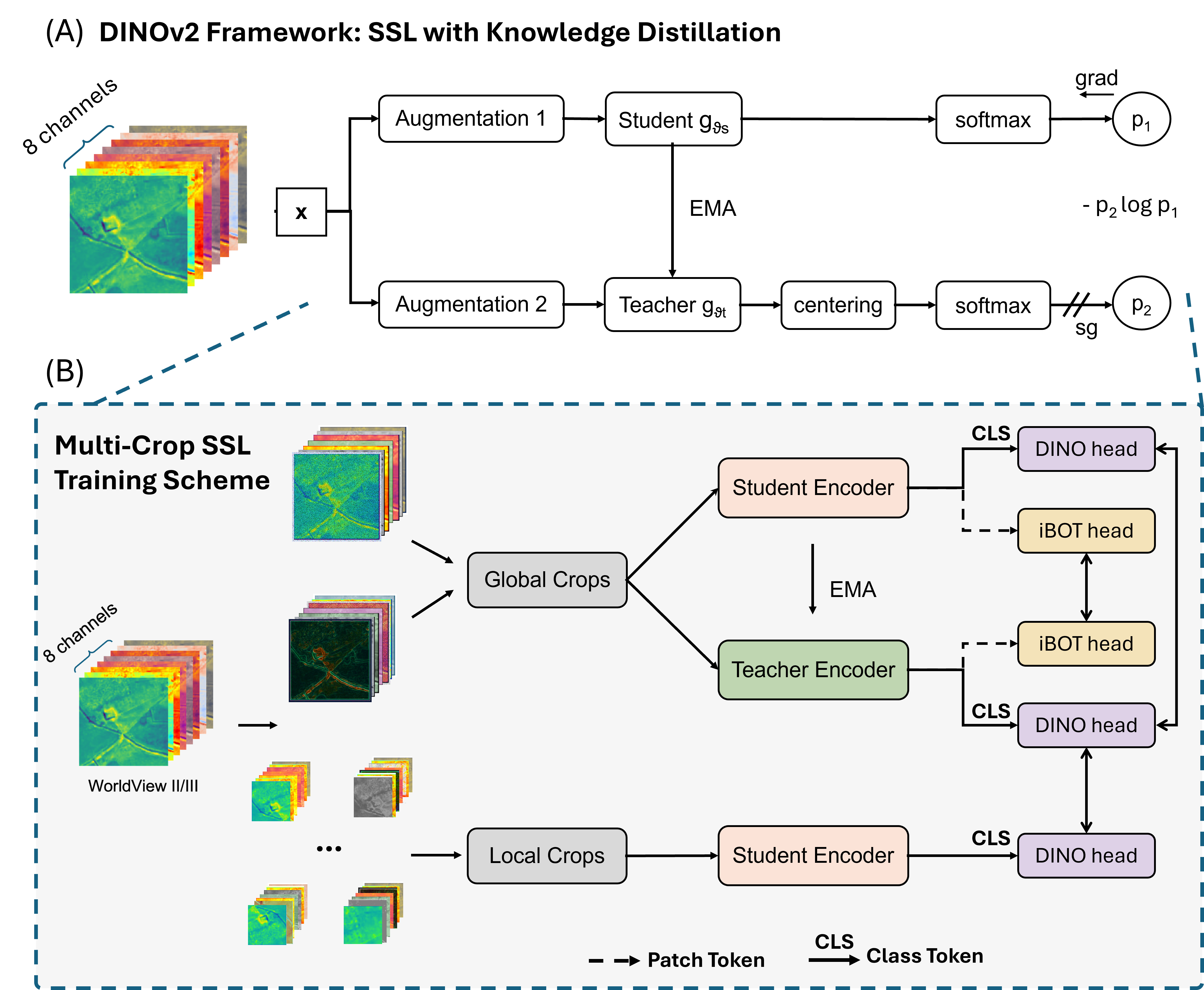}
\end{center}
\caption{\textbf{DINOv2: the self-supervised contrastive representation learning algorithm with knowledge distillation.} (A) shows an overview of the framework. (B) illustrates the details of the DINOv2 multi-crop SSL training scheme.}
{ \label{fig:dino_simple}}
\end{figure*}

\subsection{DeepAndes}
\subsubsection{\textbf{DINOv2 Architecture}}
DINOv2 (and its earlier variant DINO~\cite{DINO}) employs a contrastive learning framework with knowledge distillation, aiming to maximize the similarity of feature representations between two sets of augmented views of the same input image. Figure~\ref{fig:dino_simple}A illustrates the main architecture of DINOv2. The student network and teacher network use the same network architecture g, a vision transformer, but with different weights $\theta_s$ and $\theta_t$. During pre-training, the student network  $\text{g}_{\theta_s}$ is trained to match the representation of the teacher network $\text{g}_{\theta_t}$. As shown, for an input image $x$, two sets of augmented views are generated and sent to $\text{g}_{\theta_s}$ and $\text{g}_{\theta_t}$. Compared to $\text{g}_{\theta_t}$, the student network is information-limited because the augmented views sent to $\text{g}_{\theta_s}$ are more noisy \cite{DINO}. Unlike conventional knowledge distillation, the teacher's weights are dynamically updated using exponential moving average (EMA) of past student weights. Depending on the image views (global or local views) matched between the student and teacher, the corresponding feature representations (class tokens and patch tokens) are further projected to obtain image-level and patch-level cross-entropy loss objectives. More details on the model design and derivation can be found in \cite{DINO, oquab2023dinov2}.

\subsubsection{\textbf{SSL Pre-training}}
Similar to \cite{vorontsov2023virchow}, Figure~\ref{fig:dino_simple}B shows the detail of the training paradigm of DINOv2-based SSL pre-training. As shown in Figure~\ref{fig:dino_simple}B, two sets of augmented views are sent to the student and teacher, respectively. The student network receives both global and local crops as inputs, while the teacher network is only provided with global crops. Specifically, The global crop encompasses most of the original image, whereas the local crop focuses on a small portion of it. As shown, during training, different augmented crops of the same image are fed into the networks to obtain the image-level and patch-level loss objectives.
\begin{itemize}
    \item \textbf{DINO Loss: Image-level objective.} For global crops, the student attempts to generate the class token to match the teacher class token. The local crops are fed only to the student which tries to produce a representation that matches the teacher class token generated from the global crops, helping the model learn the local-global representations. 

    \item \textbf{iBOT Loss: Patch-level objective.} The global crops are randomly masked given to the student network. iBOT head is applied to match the (unmasked) teacher patch tokens to the student's corresponding masked patch tokens, which encourages learning fine-grained patch-level feature representations.
\end{itemize}

\textcolor{black}{In our work, we selected the ViT ``large'' (ViT-L/14) with 307M parameters as the backbone. The default multi-crop DINOv2 training uses a scale factor $s = 0.32$ to define the scaling range of local crop as (0.05, $s$), and ($s$, 1) for global views. In our study, we applied the scaling range (0.2, 0.5) for the local view and (0.5, 1) for the global view, due to the sparse image feature distribution and noise present in our remote sensing images. The other default hyperparameters for DINOv2 training algorithm were used for DeepAndes, as detailed in \cite{oquab2023dinov2} with the following modifications: a base learning rate of 0.0002, and the entire training process covering approximately 40 passes over the 3-million-image dataset. DeepAndes was trained using AdamW ($\beta_1 = 0.9, \beta_2 = 0.999$) with float16 precision. For ViT-L, we used 65,536 prototypes, resulting in 65,536-dimensional projection heads. Distributed data parallel (DDP) training was employed across 8 NVIDIA DGX-A100 nodes, with a batch size of 64 per GPU node. To implement complex augmentation pipelines, we adapted the default augmentation modules for 8-band image data with minor modifications. \textcolor{black}{The summary of hyperparameters for pre-training 3 million image crops is provided in Table~\ref{tab:ssl_settings}.}}

\subsubsection{\textbf{Feature Embeddings}}
For an input image of size 256 $\times$ 256, it is resized to 224 $\times$ 224 before being input to the model, in accordance with the default settings of DINOv2. Using ViT-L with a patch size of 14 $\times$ 14, the pre-trained DeepAndes projects the input image into vector representations (tokens) of dimension 1,024 in the feature space. With the default settings, this includes one class token and 256 image patch tokens. These vectors generated by foundation models present meaningful feature information of the input images with a reduced dimension can be further utilized for diverse downstream tasks \cite{vorontsov2023virchow, DINO-MC, guo2024skysense}. In our case, the class token can be used for downstream tasks such as image classification by concatenating a simple linear classifier. The patch tokens can be utilized for pixel-level image recognition tasks, such as semantic segmentation, by connecting them to a segmentation head.

\begin{table*}[ht]
\centering
\caption{Summary of pre-training hyperparameters. Other settings are default as~\cite{oquab2023dinov2}}
\renewcommand{\arraystretch}{1.5} 
\begin{adjustbox}{max width=\textwidth}
\begin{tabular}{p{3.5 cm} p{15.5cm}}
\toprule
\textbf{Category} & \textbf{Key Settings} \\
\midrule
\textbf{Architecture} & ViT-Large; Input channels = 8; Patch size = 14; 0.3B parameters. \\
\textbf{Student} & Drop path rate = 0.3; Layerscale = 1e-5; Drop path uniform = true; FFN layer = MLP; 
$qkv$, projection, and FFN bias = true; No register tokens; Interpolate offset = 0.1. \\
\textbf{Teacher} & 
Momentum = 0.992 $\rightarrow$ 1.0; temperature = 0.04 $\rightarrow$ 0.07; Warmup epochs = 30. \\
\textbf{Training Setup} & 
Epochs = 500; Iterations per epoch = 1000; Batch size per GPU = 64; GPUs = 8; Warmup epochs = 80; Gradient clip = 3.0. \\
\textbf{Optimization} & 
Base LR = 2e-4 (scaled $\sqrt{1024/\text{batch}}$); Weight decay = 0.04 $\rightarrow$ 0.4; Layerwise decay = 0.9; AdamW $(\beta_1,\beta_2)=(0.9,0.999)$. \\
\textbf{Loss / Heads} & 
DINO and iBOT loss weights = 1.0 each; iBOT mask ratio = 0.1–0.5; Separate iBOT head = true; Prototypes = 65,536; Head bottleneck dim = 256; Hidden dim = 2,048; Layers = 3. \\
\textbf{Global-Local Crops} & 
Global crops: size = 224, scale = 0.5–1.0; Local crops: size = 98 (divisible by 14), number = 8, scale = 0.2–0.5. \\
\textbf{Mixed precision} & fp16 for parameters and reduce; fp32 for buffers \\
\bottomrule
\end{tabular}
\end{adjustbox}
\label{tab:ssl_settings}
\end{table*}

\subsection{Archaeological Downstream Analysis}

\begin{figure*} [hb]
\begin{center}
\includegraphics[width=0.9\linewidth]{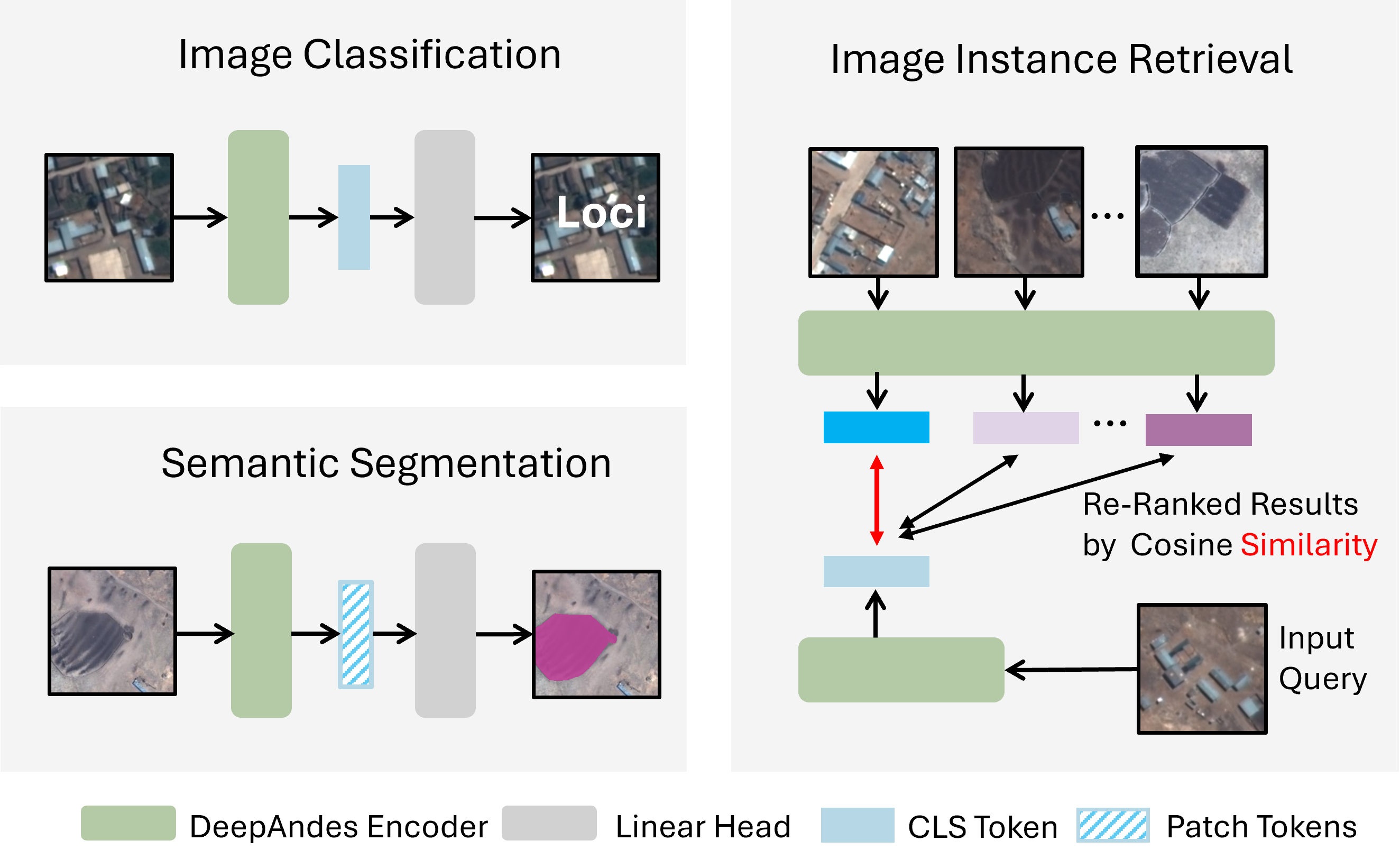}
\end{center}
\caption{\textbf{Image Understanding Downstream Analysis.} Both image-level tasks (e.g., image classification and retrieval) and pixel-level tasks (e.g., segmentation) are included.}
{ \label{fig:downstream_tasks}}
\end{figure*} 

\subsubsection{\textbf{Handling Imbalance in Archaeological Data}}
Another major challenge in archaeological remote sensing is the inherent imbalance of archaeological data. One of the previous studies~\cite{ssl_prior} shows that the proportion of our satellite imagery that include archaeological settlement features is typically low, at less than 7\% in the surveyed areas. For example, in a manually labeled dataset consisting of 5,830 labeled images sourced from a 4,000~km$^2$ survey area, the ratio of ``positive'' (containing settlement features) to ``negative'' (not containing settlement features) images can be as low as 1:100.

\textcolor{black}{To address the data imbalance, common approaches include data resampling or ensemble methods~\cite{leevy2018survey}, typically rely on extensive domain expertise or labeled data and are difficult to scale effectively. Foundation models leverage large-scale unlabeled data through task-agnostic pre-training, which has been shown in~\cite{hong2024spectralgpt, shi2023long} to alleviate data imbalance and reduce overfitting in downstream tasks. K-fold cross-validation was further employed to validate the robustness of this approach in addressing the imbalance scenario.}

\subsubsection{\textbf{Archaeological Downstream Tasks}}
To support the motivation above, we evaluate DeepAndes on three image understanding downstream tasks in archaeology: image-level classification and retrieval, and pixel-level segmentation, as shown in Figure~\ref{fig:downstream_tasks}.

\textbf{Imbalanced Loci Classification.} Loci classification, or archaeological data interest classification, is the process of distinguishing image crops with high archaeological relevance from those that are less informative. This step is crucial in large-scale remote sensing surveys, where vast amounts of satellite imagery include diverse content—natural landforms, vegetation, and modern infrastructure—of which only a small fraction is archaeologically significant. Effective classification of Andean corrals and buildings directs surveyors to the most relevant imagery, saving their time and improving archaeological interpretation. Specifically, images in the classification dataset were labeled as either ``positive” (loci present) or ``negative” (no presence of loci). The class token of the pre-trained ViT-L backbone was connected to a linear classifier with two fully connected layers to predict these classes. To reflect data imbalance, the positive-to-negative ratio was set to 1:10. Few-shot performance was then evaluated at different dataset scales (10\%, 30\%, 50\%, and 100\%).

\textbf{Image Instance Retrieval}. Another image-level downstream task conducted in this study is image instance retrieval. Unlike classification, image retrieval does not require fine-tuning on DeepAndes. Instead, we use an image of archaeological loci as the input query and retrieve the top-k most similar images from the database, ranking them based on their cosine similarity to the query loci image (via the class token). Image retrieval enables rapid construction of relevant datasets without additional supervised training. With only a small set of labeled loci images, we can expand surveyed datasets by retrieving semantically similar samples from unexplored imagery. For example, Figure~\ref{fig:downstream_tasks} shows a query image with modern buildings and active corrals retrieving samples with comparable features.

\textbf{Few-shot Locus Segmentation.} As shown in Figure~\ref{fig:downstream_tasks}, we also evaluate semantic segmentation as a downstream task to recognize dense features. The goal is to delineate archaeologically relevant features within an image crop—such as corrals and buildings—while ignoring background elements. This provides pixel-level localization for analyzing settlement patterns and land use. Given the scarcity of labeled archaeological data, we adopt few-shot settings (e.g., 10-30 training samples) to evaluate whether the pre-trained model can adapt to segmentation with minimal annotation. Specifically, we perform few-shot semantic segmentation tasks across three challenging loci: active buildings, active corrals, and archaeological corrals. Similar to~\cite{oquab2023dinov2}, patch-level features are extracted from the images via patch tokens and then concatenated with a simple linear segmentation head to generate output logits.

\section{Data and Experiments}

\subsection{Data collection and Pre-processing}
\textcolor{black}{\textbf{Data collection.}} The remote sensing satellite images used in this work were collected by the WorldView-2 and WorldView-3 satellites between 2010 and 2023, and imagery was provided by Maxar through the NextView Imagery License agreement with the National Geospatial-Intelligence Agency (NGA). We excluded images that are heavily obscured by clouds, as the underlying content is scarcely visible and therefore unsuitable for both machine learning analysis and manual survey. We observed that archaeological features—such as buildings, corrals, and related structures—are generally more discernible in imagery acquired during the dry season, whereas vegetation growth in the wet season often obscures these features. Consequently, whenever possible, we prioritize dry-season imagery (June through October). However, in cases where only wet-season data are available, we also include those images to ensure broader coverage.

\textbf{Data pre-processing.} First, the collected satellite images underwent color correction and orthographic correction using a coarse digital elevation model (DEM). The data are then pan-sharpened using the Bayesian fusion algorithm from Orfeo-Toolbox~\cite{grizonnet2017orfeo} to increase the spatial resolution of the multi-spectral imagery to 0.31 m for WorldView-2 imagery to match the native resolution of WorldView-3 imagery. In this work, all 8 spectral bands (four standard colors—red, green, blue, and near-infrared 1—and four new bands—coastal, yellow, red edge, and near-infrared 2) are used. Lastly, the imagery is re-sampled from 32 bits to 8 bits to reduce storage size and computational load.

\subsection{SSL with Different Pre-training Scales}
As introduced in Section \ref{methods:train_dataset}, after data pre-processing, we construct the self-supervised pre-training dataset sourced from the entire Peruvian Andes. In total, the imagery covers survey regions of approximately 488,640 $km^{2}$ and includes 8 distinct land cover types. We densely sample 3 million image patches of size 256 $\times$ 256 from all surveyed regions to ensure a diverse, large-scale pre-training dataset. 

\textcolor{black}{In this work, we perform SSL using pre-training datasets of varying sizes to robustly evaluate the impact of DINOv2-based pre-training on downstream tasks. Specifically, we pre-train the DeepAndes backbone on datasets containing 30K, 300K, and 3 million image crops, denoted as \textbf{FM30K}, \textbf{FM300K}, and \textbf{FM3M}, respectively. Since our input data consists of 8-band imagery rather than standard RGB images, we also include a \textbf{Scratch model}—identical in architecture to the DeepAndes backbones (FM30K, FM300K, and FM3M) but initialized with random weights, i.e., without any pre-training—as a baseline for comparison.}

\textcolor{black}{All model backbones (FM30K, FM300K, FM3M, and Scratch) adopt the ViT-L/14 architecture, a Vision Transformer Large variant with 24 transformer layers, a hidden size of 1024, and a patch size of 14 $\times$ 14, totaling approximately 0.3 billion parameters. Input images are resized to 224 $\times$ 224 following the default configuration of the ViT-L/14 backbone in DINOv2. To accommodate 8-band imagery, the original patch embedding layer is replaced with a 2D convolutional layer that accepts 8 input channels. The ViT encoder outputs 257 tokens, including one class token and 256 patch tokens, each with a hidden dimension of 1,024.}

\subsection{Archaeological Downstream Tasks}

\subsubsection{\textbf{Imbalanced Loci Classification}}
For the archaeological loci classification task, the dataset is imbalanced, with a 1:10 ratio of ``positive'' (containing loci) to ``negative'' (not containing loci) images. Due to this imbalance and limited labeled data, we employ K-fold cross-validation (K=5) for robust evaluation. Specifically, for each train-test split, four folds (positive: 729, negative: 7,290) are used for training, and one fold (positive: 183, negative: 1,830) is used for testing. A simple linear classifier with two fully connected layers is concatenated to the DeepAndes backbone. Prior to the five-fold cross-validation experiments, we ran several random seed trials to identify optimized hyperparameters for model fine-tuning. Additionally, to evaluate the few-shot learning capability of the pre-trained transformer backbones, we assess the classification task at different downstream dataset scales (10\%, 30\%, 50\%, and 100\% of the original classification dataset).

\textbf{Evaluation Metrics.} 
\textcolor{black}{
Given the imbalanced nature of the classification task, we evaluate model performance using Precision, Recall, F1 score, and the area under the precision--recall curve (PR-AUC). Particularly, Precision is defined as the proportion of predicted positives that are correct. This is calculated as the ratio of true positives (TP) to the sum of TP and false positives (FP):
\[
Precision = \frac{TP}{TP + FP}
\]
The Recall (or sensitivity) measures the proportion of actual positives correctly identified. This is calculated as the ratio of TP to the the sum of TP and false negatives (FN):
\[
Recall = \frac{TP}{TP + FN}
\]
The F1 score combines both Precision and Recall into a single metric:
\[
F1\ score = 2 \times \frac{Precision \times Recall}{Precision + Recall}
\]
The PR-AUC summarizes performance across thresholds by integrating Precision as a function of Recall:
\[
PR\text{-}AUC = \int_{0}^{1} Precision(Recall)\, d(Recall)
\]
The combined use of these metrics enables a more comprehensive and fair evaluation in scenarios where the minority class (archaeological loci) is of interest.}


\subsubsection{\textbf{Image Instance Retrieval}}
For the image instance retrieval task, we use all available labeled data (both \textbf{training} and \textbf{testing}) from the loci classification task and probe the frozen backbones through the class token to form a database. This can be implemented using FAISS library \cite{FAISS}. In our database, for each positive sample, we use it as the input query and retrieve the top $k$ most similar instances from the database, ranking them based on their cosine similarity to the query image. The precision of retrieving image instances from the same class as the query image (positive samples) is then evaluated for different pre-trained backbones.

\textbf{Evaluation Metric.} To evaluate the performance of image instance retrieval task, we use mAP@$k$ (mean Average Precision within top-\textit{k} retrieved samples). The derivation is following. When performing image retrieval, we rank the images based on their similarity to the query image, where rank 1 corresponds to the most similar image, rank 2 to the second most similar image, and so on. The precision at a given rank is defined as the proportion of relevant images (those belonging to the same class as the query image) retrieved up to that rank:

\[
Precision@i \ = \frac{\text{Number of images retrieved at rank } i}{i}
\]
Then, the Average Precision (AP) for a given query is the average of precision values at different ranks (up to \textbf{\textit{k}}) where relevant images are retrieved:

\[
AP = \frac{1}{N}\sum_{i=1}^{N}Precision@i
\]
where \( N \) is the number of relevant images for that query. Finally, mAP@$k$ is calculated by averaging the AP over all queries in the dataset:

\[
mAP@k = \frac{1}{Q} \sum_{q=1}^{Q} AP_q
\]
where \( Q \) is the total number of queries (number of positive samples in the database). In this work, we use multiple \textbf{\textit{k}} values for evaluation, ranging from 5, 20, and 50, to 100.

\subsubsection{\textbf{Few-shot Locus Segmentation}}
To evaluate the foundation model's few-shot learning performance on pixel-level feature recognition, we focus on the semantic segmentation of three specific types of loci---active buildings, active corrals, and archaeological corrals. Three small-scale binary semantic segmentation datasets are constructed: the Active Buildings dataset contains 48 images, the Active Corrals dataset contains 55 images, and the Archaeological Corrals dataset contains 46 images. K-fold cross-validation (K=5) is employed for robust evaluation. The segmentation head is simple and directly utilizes the learned features from pre-training. Specifically, patch-level features are extracted from the images using patch tokens and concatenated with a simple one-layer linear segmentation head. Prior to the five-fold cross-validation experiments, we conducted several random seed trials to identify optimized hyperparameters for model training. Similar to the few-shot loci classification, we also evaluate the pre-trained model's segmentation performance at different downstream dataset scales, including 10, 20, and 30 images.

\textbf{Evaluation Metric.} 
The Dice Similarity Coefficient (DSC) is used to evaluate the precision of loci segmentation in this work. This metric ensures precise pixel-level overlap between the predicted and ground truth masks. It emphasizes the importance of accurate segmentation in smaller, localized areas, which aligns with our segmentation datasets, where loci objects (foreground) are typically much smaller than the background areas.

\[
DSC = \frac{2 \times TP}{2 \times TP + FP + FN}
\]

\begin{itemize}
    \item \( TP \) represents the number of pixels correctly classified as the loci in both the predicted and ground truth masks.
    \item \( FP \) represents the number of pixels incorrectly classified as loci in the predicted mask but not in the ground truth mask.
    \item \( FN \) represents the number of pixels incorrectly classified as background in the predicted mask but actually belonging to the loci in the ground truth mask.
\end{itemize}

\section{Results}

\begin{table*}[ht]
\centering
\renewcommand{\arraystretch}{2} 
\begin{adjustbox}{max width=\textwidth}
\begin{tabular}{|c|c c c|c c c|c c c|c c c|}
\hline
\multirow{2}{*}{Model} & \multicolumn{3}{c|}{\(N_{\text{train}} = 729\)} & \multicolumn{3}{c|}{\(N_{\text{train}} = 365\)} & \multicolumn{3}{c|}{\(N_{\text{train}} = 218\)} & \multicolumn{3}{c|}{\(N_{\text{train}} = 72\)} \\
\cline{2-13} 
 & F1 & Recall & Precision & F1 & Recall & Precision & F1 & Recall & Precision & F1 & Recall & Precision \\
\hline
Scratch & 0.544 & 0.457 & 0.735 & 0.402 & 0.288 & 0.692 & 0.26 & 0.241 & 0.52 & -- & -- & -- \\
FM30K & 0.654 & 0.595 & 0.747 & 0.596 & 0.522 & 0.701 & 0.516 & 0.434 & 0.672 & 0.418 & 0.342 & 0.556 \\
FM300K & 0.824 & 0.806 & 0.848 & 0.804 & 0.795 & 0.817 & 0.788 & 0.782 & 0.792 & 0.728 & 0.671 & 0.812 \\
FM3M &  \textbf{0.886} & \textbf{0.876} & \textbf{0.894} & \textbf{0.872} & \textbf{0.872} & \textbf{0.875} & \textbf{0.862} & \textbf{0.866} & \textbf{0.857} & \textbf{0.830} & \textbf{0.825} & \textbf{0.837} \\
\hline
\end{tabular}
\end{adjustbox}
\caption{\textbf{Performance on Imbalanced Loci Classification.} The mean F1 scores, Recall (R), and Precision (P) from the five-fold cross-validation are presented. We compare performance of four model backbones: ViT-L trained from scratch, and DeepAndes pre-trained using 30K, 300K, and 3M images. \textbf{$N_{\text{train}}$} represents the scale of \textbf{``positive''} images (containing loci) in the training dataset. The \textbf{``positive'' to ``negative''} ratio is \textbf{1:10} in both training and testing set. Entries marked with ``--'' indicate that the experiments do not converge or the values are not supported. For clarity, the highest metric values of each few-shot dataset evaluation are highlighted in \textbf{bold}.}
\label{tab:classification_performance}
\end{table*}

\subsection{Imbalanced Loci Classification}
\begin{figure*} [ht]
\begin{center}
\includegraphics[width=1\linewidth]{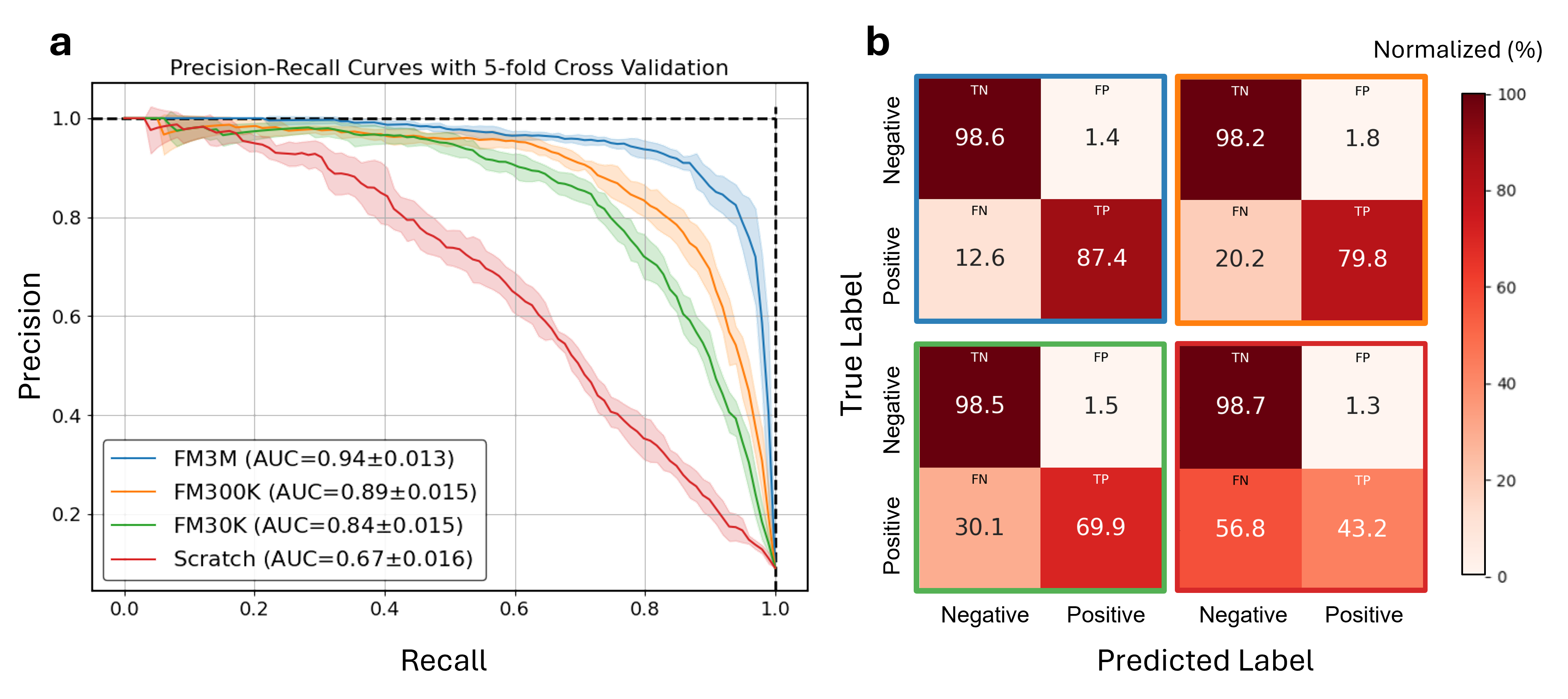}
\end{center}
\caption{\textbf{Performance on Imbalanced Loci Classification:} Precision-Recall curves from five-fold cross-validation \textbf{(a)}. Confusion matrices (normalized) for each model shown for a representative hold-out fold \textbf{(b)}.}
{ \label{fig:classification_plot}}
\end{figure*}

Both Table~\ref{tab:classification_performance} and Figure~\ref{fig:classification_plot} display the five-fold cross-validation results for imbalanced loci classification. Specifically, Table~\ref{tab:classification_performance} details the mean F1 scores, Precision (Prec), and Recall (Rec). Figure~\ref{fig:classification_plot}a illustrates the Precision-Recall (PR) curves, while Figure~\ref{fig:classification_plot}b presents the confusion matrices for each model, highlighted using a representative hold-out fold. 

As shown in Table~\ref{tab:classification_performance}, the Scratch model struggles with limited and imbalanced data, converging only at \( N_{train} = 729 \) (F1 = 0.544) and \( N_{train} = 365 \) (F1 = 0.402), while failing entirely at smaller datasets. In contrast, DeepAndes models with SSL pre-training improve consistently across all dataset sizes. At \( N_{train} = 72 \), FM3M achieves F1 = 0.83, Recall = 0.825, and Precision = 0.837, effectively balancing false positives and false negatives. Smaller pre-trained models decline in performance, with FM300K dropping to F1 = 0.728, Rec = 0.671, Prec = 0.812 and FM30K to F1 = 0.418, Rec = 0.342, Prec = 0.556, reflecting higher false negatives due to class imbalance. The same pattern can also be observed in Figure~\ref{fig:classification_plot}a and b, where the highlighted blue PR curve has the highest PR-AUC and seemingly fewer false negatives compared to the other models. 

Additionally, as Table.~\ref{tab:classification_performance} indicated, FM3M maintains strong results with few labels; at \( N_{train} = 72 \), the larger pre-trained FM3M emerges as a better few-shot learner, achieving performance comparable to FM300K at \( N_{train} = 729 \). These results underscore the advantages of large-scale SSL pre-training in managing imbalanced data and enhancing few-shot learning.

\subsection{Image Instance Retrieval}
\begin{figure*} [!htbp]
\begin{center}
\includegraphics[width=0.9\linewidth]{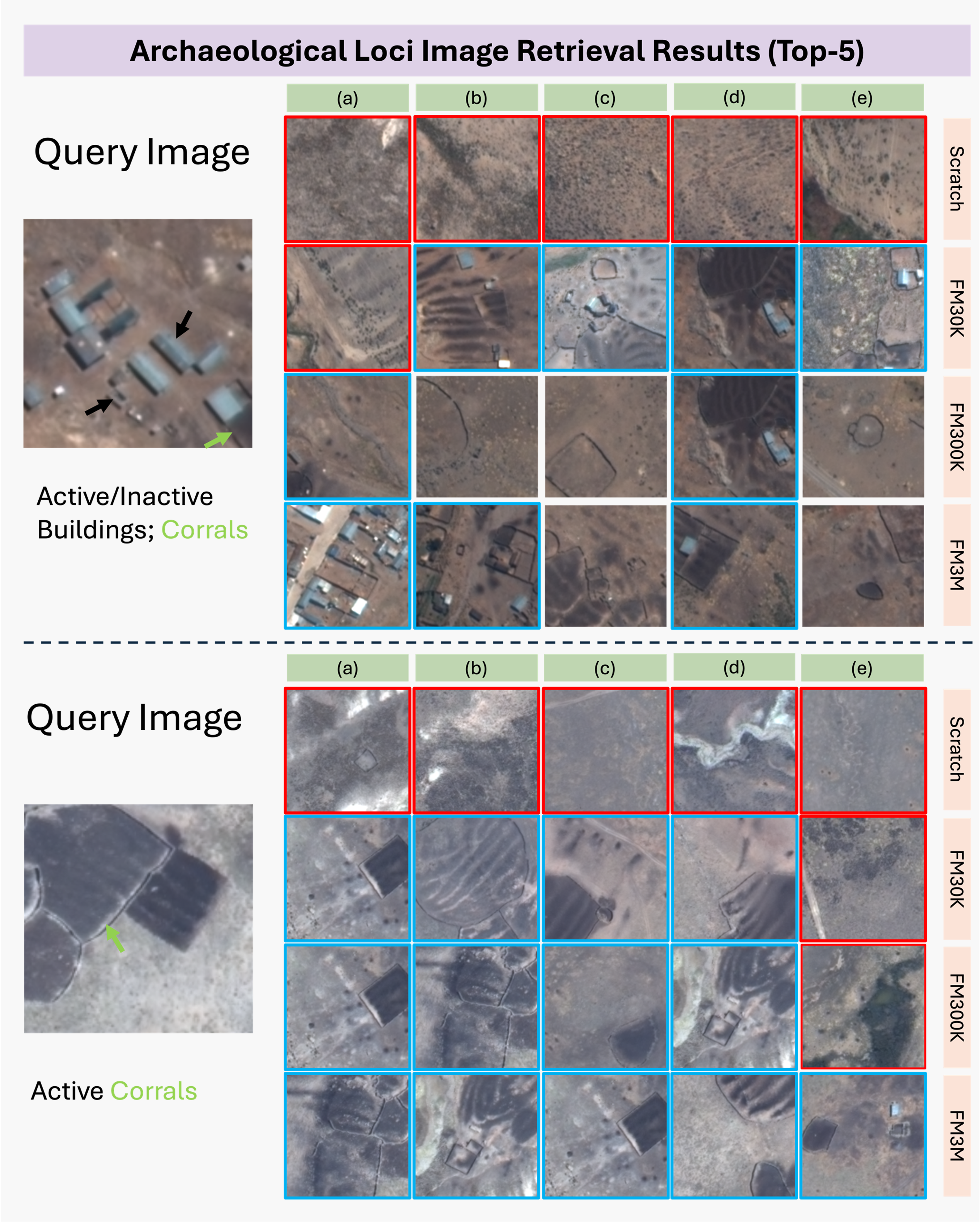}
\end{center}
\caption{\textbf{Examples of Retrieved Images from Different Pre-trained Models.} The left column displays two example query images, with arrows pointing to key archaeological features. Both query images are from the positive class in the imbalanced loci classification dataset. The right column presents the top-5 retrieved images based on cosine similarity. The images with a \textcolor{red}{red} box indicate incorrect class retrieval, while the \textcolor{blue}{blue} box highlights correct retrieval of both the image class and all archaeological features in the query image.}
{ \label{fig:img_retrieval_plot}}
\end{figure*}

For loci image instance retrieval task, the performance of four frozen ViT-L backbones (Scratch, FM30K, FM300K, and FM3M) is compared in Table \ref{tab:retrival_performance}. As demonstrated, for the image retrieval task, larger $k$ values result in lower mAP@\textit{k} scores, indicating that retrieving more samples from the database also introduces more irrelevant ones relative to the query image class. It is also evident that pre-training improves retrieval performance. The Scratch model, without any pre-training, achieves the lowest mAP@\textit{k} scores across all choices of \textit{k} for this evaluation, with mAP@\textit{5} starting at 0.444. In contrast, pre-trained DeepAndes models—FM30K, FM300K, and FM3M—show progressively higher mAP values, with FM3M achieving the highest performance, with mAP@\textit{5} of 0.869. Figure~\ref{fig:img_retrieval_plot} provides qualitative visualizations of the retrieved examples. As pre-training scales up, FM3M retrieves more relevant images (highlighted in blue boxes) that correctly match both the image class and features of the query image. In contrast, FM30K and the Scratch model retrieve incorrect image classes (highlighted in red boxes) among the top-5 retrieved examples. These results clearly demonstrate that SSL pre-training, particularly at a larger scale, enhances image-level feature representations and improves instance retrieval accuracy.

\begin{table}[ht]
\renewcommand{\arraystretch}{2} 
\centering
\setlength{\tabcolsep}{2mm}
\begin{tabular}{c c c c c}
\hline
\textbf{Model} & \textbf{mAP@\textit{5}} & \textbf{mAP@\textit{20}} & \textbf{mAP@\textit{50}} & \textbf{mAP@\textit{100}} \\ \hline
Scratch & 0.444 & 0.391 & 0.330 & 0.296 \\ 
FM30K & 0.498 & 0.435 & 0.384 & 0.350 \\ 
FM300K & 0.757 & 0.677 & 0.597 & 0.532 \\ 
FM3M & \textbf{0.869} & \textbf{0.804} & \textbf{0.744} & \textbf{0.690} \\ \hline
\end{tabular}
\caption{\textbf{Performance on Image Instance Retrieval.} The mAP@$k$ (mean Average Precision within top-$k$ retrieved samples) for different choices of $k$ (5, 20, 50, and 100) is presented. Four frozen ViT-L backbones are compared: ViT-L (scratch) and DeepAndes pre-trained using 30K, 300K, and 3M images. For clarity, the highest metric values are highlighted in \textbf{bold}.}
\label{tab:retrival_performance}
\end{table}

\subsection{Few-shot Locus Segmentation}

\begin{table*}[ht]
\centering
\Large
\renewcommand{\arraystretch}{1.5} 
\begin{adjustbox}{width=1\textwidth}
\begin{tabular}{|c|c|c c c c|c c c c|c c c c|}
\hline
\multirow{2}{*}{Backbone} & \multirow{2}{*}{Model} & \multicolumn{4}{c|}{Active Buildings\textsuperscript{1}} & \multicolumn{4}{c|}{Active Corrals\textsuperscript{2}} & \multicolumn{4}{c|}{Archaeol. Corrals\textsuperscript{3}} \\ \cline{3-14}
 & & 100\% & \(N_{\text{train}} = 10\) & \(N_{\text{train}} = 20\) & \(N_{\text{train}} = 30\) & 100\% & \(N_{\text{train}} = 10\) & \(N_{\text{train}} = 20\) & \(N_{\text{train}} = 30\) & 100\% & \(N_{\text{train}} = 10\) & \(N_{\text{train}} = 20\) & \(N_{\text{train}} = 30\) \\ \hline
\multirow{4}{*}{Frozen} & Scratch & 29.1 & 8.2 & 21.3 & 20.6 & 38.3 & 11.6 & 14.6 & 21.0 & 8.4 & -- & -- & -- \\
 & FM30K & 27.0 & 20.0 & 23.0 & 24.7 & 45.7 & 45.4 & 44.8 & 45.5 & 11.9 & 9.7 & 11.9 & 12.1 \\
 & FM300K & 55.2 & 41.4 & 47.3 & 51.4 & 60.9 & 55.5 & 56.5 & 59.3 & 27.8 & 19.0 & 17.8 & 24.8 \\
 & FM3M & 58.3 & \underline{52.3} & 56.2 & 58.2 & \underline{70.4} & \underline{65.5} & \underline{67.1} & \underline{68.2} & \underline{67.3} & \textbf{56.3} & \underline{61.3} & \underline{63.0} \\ \hline
\multirow{4}{*}{Finetuned} & Scratch & 47.1 & 42.1 & 47.8 & 45.9 & 64.2 & 57.9 & 62.0 & 62.5 & 11.6 & 15.1 & 15.4 & 15.4 \\
 & FM30K & 53.8 & 44.0 & 52.8 & 53.2 & 66.4 & 61.3 & 63.7 & 64.8 & 17.7 & 16.2 & 15.6 & 18.4 \\
 & FM300K & \underline{59.7} & 51.4 & \underline{57.2} & \underline{58.8} & 69.2 & 59.3 & 64.6 & 67.8 & 33.9 & 15.3 & 17.9 & 29.4 \\
 & FM3M & \textbf{69.3} & \textbf{57.8} & \textbf{63.7} & \textbf{66.5} & \textbf{81.1} & \textbf{70.8} & \textbf{73.5} & \textbf{75.5} & \textbf{84.8} & \underline{51.3} & \textbf{72.5} & \textbf{81.0} \\ \hline
\end{tabular}
\end{adjustbox}

\vspace{0.5em}
\begin{minipage}{\textwidth}
\footnotesize
\raggedright
\textsuperscript{1} \textbf{Active Buildings Dataset} contains \textbf{48} images featuring modern buildings against dense forest or vegetation backgrounds.\\
\textsuperscript{2} \textbf{Active Corrals Dataset} contains \textbf{55} images, each showing corrals with visible evidence of animal use.\\
\textsuperscript{3} \textbf{Archaeological Corrals Dataset} contains \textbf{46} images showing corrals where signs of use have disappeared.
\end{minipage}

\caption{\textbf{Performance on Few-shot Segmentation of Three Loci Types.} Mean DSC scores from five-fold cross-validation are presented, which include both transfer learning (frozen backbones) and fine-tuning (unfrozen backbones) across four models: Scratch, FM{30K}, FM{300K}, and FM{3M}. Specifically, $N_\textbf{train}$ indicates the scale of the training dataset. \textbf{Bold} entries denote the highest DSC score for each few-shot setting, while \textbf{\underline{underscored}} entries indicate the second highest. Entries marked with \textbf{``--''} indicate that the experiments do not converge or the values are not supported.}
\label{tab:segmentation}
\end{table*}

For few-shot locus segmentation, Table \ref{tab:segmentation} summarizes the models' performance on three locus types (i.e., active buildings, active corrals, and archaeological corrals). It is evident that as the pre-training scales increase—from 30K to 300K, and then to 3M—the model's performance in both transfer learning (frozen) and fine-tuning improves compared to training from scratch. As highlighted in Table \ref{tab:segmentation}, the best model performance across all few-shot datasets and tasks comes from the FM3M. For \textbf{active/archaeological corrals} segmentation tasks, simply training the linear segmentation head on top of the frozen FM{3M} backbone achieves the second highest DSC score across all experiments, demonstrating the benefits of our million-scale pre-training. On the other hand, models trained from scratch exhibit relatively low DSC scores, especially with small datasets. For example, at \( N_{train} = 10 \), the DSC score for the Scratch model (frozen) on the \textbf{Active Corrals} dataset is only 11.6. In contrast, frozen pre-trained models show an improvement of over 30\% starting from FM30K (45.4), with DSC scores continuing to rise as the pre-training scale increases. Although model performance improves with fine-tuning the entire ViT-L, transfer learning on the frozen FM3M backbone still exhibits DSC scores that are either comparable to or surpass those of models with smaller pre-training scales. This improvement is particularly noticeable when downstream data is very limited (e.g., \( N_{train} = 10 \)) across three segmentation tasks. Overall, these results highlight the effectiveness of million-scale DeepAndes in few-shot learning tasks, where the model can generalize well with limited labeled data. This makes it especially valuable in fields like archaeology, where data annotation is scarce.

\begin{figure*} [!htbp]
\begin{center}
\includegraphics[width=0.9\linewidth]{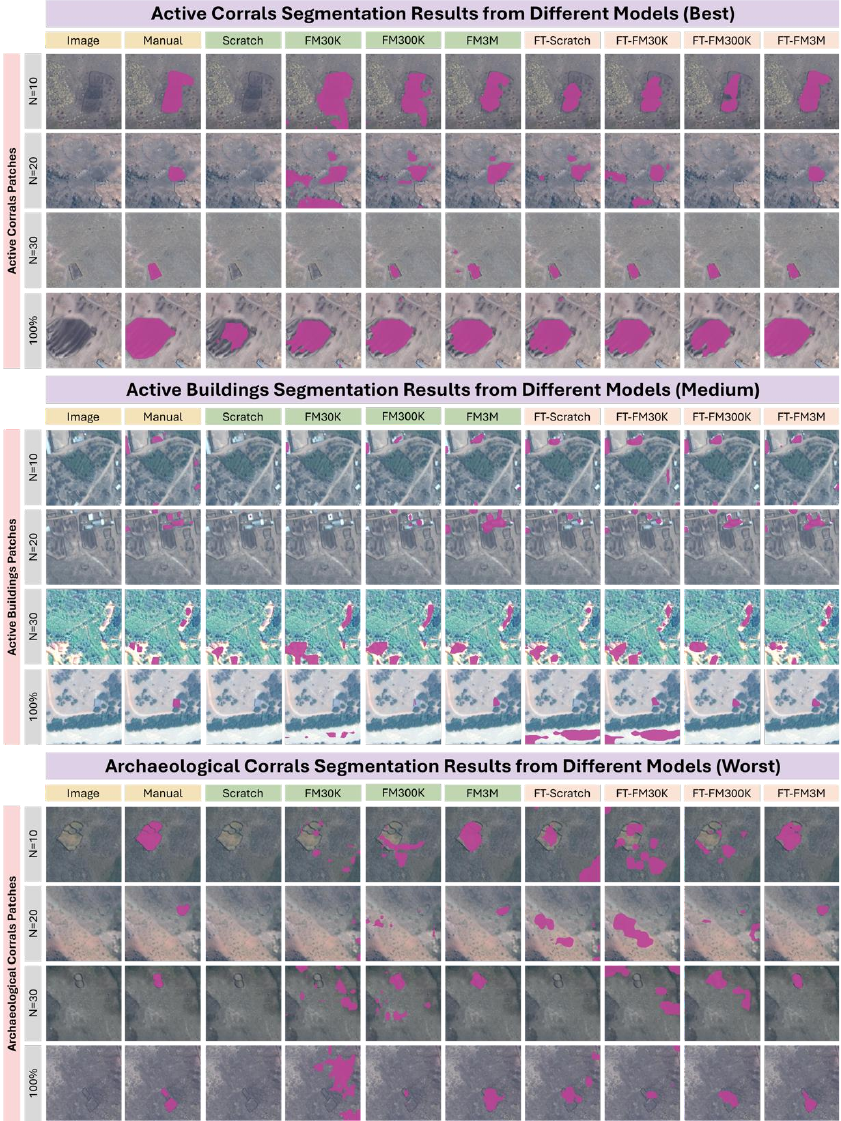}

\end{center}
\caption{\textbf{Qualitative Results of Loci Segmentation.} ``FT-'' denotes fine-tuning. For convenience, image patches are shown in RGB. Superior performance are from either FT-FM3M or FM3M.}
{ \label{fig:seg1. seg2. seg3}}
\end{figure*} 

Additionally, we present the qualitative visualizations, paired with the Table \ref{tab:segmentation} results, for the three types of loci segmentation tasks in Figure~\ref{fig:seg1. seg2. seg3}. An example patch and its segmentation results from different models are shown. As shown, segmenting archaeological corrals (bottom panel) is the most challenging, as there is little to no difference between the inside and outside, except for the boundary pixels. Segmenting active corrals (top panel) is the easiest, as visual residues of active use (e.g., removal of vegetation through grazing) creates high contrast between the interior and exterior of corrals. Active building segmentation (middle panel) shows medium performance generally. The variability in building colors and their small size makes it difficult for the model to converge. As shown, under few-shot settings (\( N_{train} = 10,\ 20,\ 30 \)), Scratch and FM30K struggled to achieve reliable performance, as evidenced by the transfer learning results for even the best-performing task (top panel). 

To some extent, fine-tuning the entire model, though more costly, helps address these issues. For example, FT-FM3M, which uses the DeepAndes (3M) backbone, effectively captures foreground pixels and, with a simple linear segmentation head, achieves the highest segmentation quality across the three tasks. However, in active building segmentation (middle panel), transfer learning with FM3M still achieves strong performance, outperforming other frozen backbones and demonstrating results comparable to the more costly full-model fine-tuning approach (e.g., FT-FM300K). For the most challenging task (bottom panel), archaeological corral segmentation, FM3M and FT-FM3M clearly outperform other models, which struggle to capture the foreground pixels and produce many false positives.

\subsection{Ablation Study}

\subsubsection{\textbf{Performance Comparison}}
As shown in Table~\ref{tab:ablation}, we benchmark our DeepAndes backbone (FM3M) against several representative baselines: a randomly initialized scratch model, SSL backbones including MAE and MoCo-V2, and SatMAE, a domain-specific remote sensing foundation model. Performance is evaluated across three selected tasks introduced previously—zero-shot image retrieval (Top-5 and Top-100 mAP), few-shot classification (F1, Recall, and Precision), and few-shot segmentation (DSC). For classification and segmentation, we report results for both the full training set and a severely constrained few-shot setting (\( N_{train} = 72 \) for classification and  \( N_{train} = 10 \)  for segmentation) to simulate real-world low-resource conditions. The best performance for each metric is highlighted. 

Across all tasks, DeepAndes consistently outperforms the baselines, highlighting the benefits of large-scale DINOv2-based pre-training on 8-band imagery. In zero-shot image retrieval, it achieves the highest Top-5 and Top-100 mAP scores (0.869 and 0.69), showing the strong generalization capability of DeepAndes without task-specific fine-tuning. 

In the few-shot classification task, models with SSL pre-training outperform the scratch model, maintaining relatively high F1, precision, and recall scores. Notably, the other remote sensing foundation model, such as SatMAE, performs worse than DeepAndes, suggesting that pre-training on different satellite sources introduces a domain gap. DeepAndes maintains a strong precision-recall balance across settings. While MAE and MoCo-V2 perform comparably in the high-data regime, their performance declines more noticeably under the limited supervision (\( N_{train} = 72 \)). 

For few-shot segmentation, DeepAndes delivers the top DSC scores in 4 out of 6 settings, specifically for first two archaeological features (active buildings and corrals). Its performance remains more consistent at \( N_{train} = 10 \), highlighting its robustness in low-data regimes. However, it is worth noting that for the segmentation of archaeological corrals, DeepAndes ranks second to MAE (0.71 vs. 0.67 and 0.66 vs. 0.56), suggesting that incorporating MAE-like strategies during pre-training may enhance the model’s ability to capture dense features.
	
\begin{table*}[ht]
\centering
\renewcommand{\arraystretch}{1.5}
\setlength{\tabcolsep}{3pt}
\begin{adjustbox}{width=\textwidth}
\begin{tabular}{lcccccccccccccc}
\toprule
\multirow{4}{*}{\textbf{Method}} 
& \multicolumn{2}{c}{\textbf{Zero-shot Retrieval}} 
& \multicolumn{6}{c}{\textbf{Few-shot Classification}} 
& \multicolumn{6}{c}{\textbf{Few-shot Segmentation} (DSC)} \\
\cmidrule(lr){2-3} \cmidrule(lr){4-9} \cmidrule(lr){10-15}
& Top-5 & Top-100 
& \multicolumn{3}{c}{\textbf{\(N_{\text{train}} = 729\)}} 
& \multicolumn{3}{c}{\textbf{\(N_{\text{train}} = 72\)}} 
& \multicolumn{2}{c}{Active Buildings} 
& \multicolumn{2}{c}{Active Corrals} 
& \multicolumn{2}{c}{Archaeol. Corrals} \\
& mAP ↑ & mAP ↑ 
& F1 ↑ & Rec ↑ & Prec ↑ & F1 ↑ & Rec ↑ & Prec ↑ 
& 100\% & \(N_{\text{train}} = 10\) & 100\% & \(N_{\text{train}} = 10\) & 100\% & \(N_{\text{train}} = 10\) \\
\midrule
Scratch   & 0.444 & 0.296 & 0.544 & 0.457 & 0.735 & --    & --    & --    & 0.291 & 0.082 & 0.383 & 0.116 & 0.084 & -- \\
MoCo-V2    & 0.511 & 0.343 & 0.88 & \textbf{0.902} & 0.864 & 0.78 & 0.727 & 0.831 & 0.184 & 0.032 & 0.559 & 0.463 & 0.328 & 0.307 \\
MAE       & 0.709 & 0.473 & \textbf{0.92} & 0.901 & \textbf{0.937} & 0.82 & \textbf{0.831} & 0.804 & 0.458 & 0.20 & 0.673 & 0.298 & \textbf{0.71} & \textbf{0.66} \\
SatMAE    & 0.578 & 0.371 & 0.87 & 0.885 & 0.853 & 0.71 & 0.70 & 0.711 & 0.051 & 0.04 & 0.163 & 0.177 & 0.026 & 0.04 \\
DeepAndes    & \textbf{0.869} & \textbf{0.69} & 0.886 & 0.876 & 0.894 & \textbf{0.83} & 0.825 & \textbf{0.837} & \textbf{0.583} & \textbf{0.523} & \textbf{0.704} & \textbf{0.654} & 0.673 & 0.563 \\
\bottomrule
\end{tabular}
\end{adjustbox}
\caption{
We benchmark DeepAndes (FM3M) against additional representative baselines: a randomly initialized Scratch model, self-supervised backbones including MoCo-V2, MAE, and SatMAE, a domain-specific remote-sensing foundation model. Models are evaluated across three selected tasks: zero-shot image retrieval (Top-5 and Top-50 mAP), few-shot classification (F1, Recall, and Precision), and few-shot segmentation (DSC). For segmentation, the backbone is frozen and only a linear segmentation head is trained. Few-shot results are reported for both the full training set and the most constrained subset (\(N_{\text{train}} = 72\) for classification and \(N_{\text{train}} = 10\) for segmentation) to simulate data-limited conditions. The highest value for each evaluation metric is highlighted in \textbf{bold}.}
\label{tab:ablation}
\end{table*}

\subsubsection{\textbf{Computational Complexity}}
Table~\ref{tab:complexity} reports the computational complexity of representative backbone models. Latency reflects the inference speed, measured as milliseconds per sample (ms/sample) on an NVIDIA RTX A6000 GPU with batch size 1 and input resolution of 224 $\times$ 224. Each value is averaged over 100 runs. FLOPs are reported in units of $1\times10^{9}$ operations (GFLOPs), and the number of parameters is given in millions (M). Transformer-based models (MAE, SatMAE, and DeepAndes) incur higher inference costs than the ResNet50 baseline (MoCo-V2). DeepAndes shows the highest FLOPs and latency, reflecting a trade-off between computational efficiency and downstream performance. This suggests further improvements on model efficiency—through compression or lightweight designs—while preserving generalization.

\begin{table*}[t]
\centering
\renewcommand{\arraystretch}{1.2}  
\begin{tabular}{lcccc}
\hline
\textbf{Method} & \textbf{Backbone} & \textbf{Latency (ms)} & \textbf{FLOPs (G)}  & \textbf{Params (M)} \\
\hline
MoCo-V2   & ResNet-50 & 2.61  & 8.65   & 23  \\
MAE       & ViT-L/16  & 7.55  & 84.15  & 304 \\
SatMAE    & ViT-L/16  & 7.83  & 84.15  & 330 \\
DeepAndes & ViT-L/14  & 10.69 & 156.29 & 304 \\
\hline
\end{tabular}
\caption{\textbf{Computational Complexity Comparison.} Latency (ms/sample) is measured on an NVIDIA RTX A6000 GPU with batch size 1 and input resolution of $224 \times 224$. Each value is averaged over 100 runs. FLOPs are reported in $1 \times 10^{9}$ operations (GFLOPs), and parameters in millions (M).}
\label{tab:complexity}
\end{table*}

\subsection{DeepAndes Training and Fine-tuning Summary}
\begin{figure*} [!htbp]
\begin{center}
\includegraphics[width=1\linewidth]{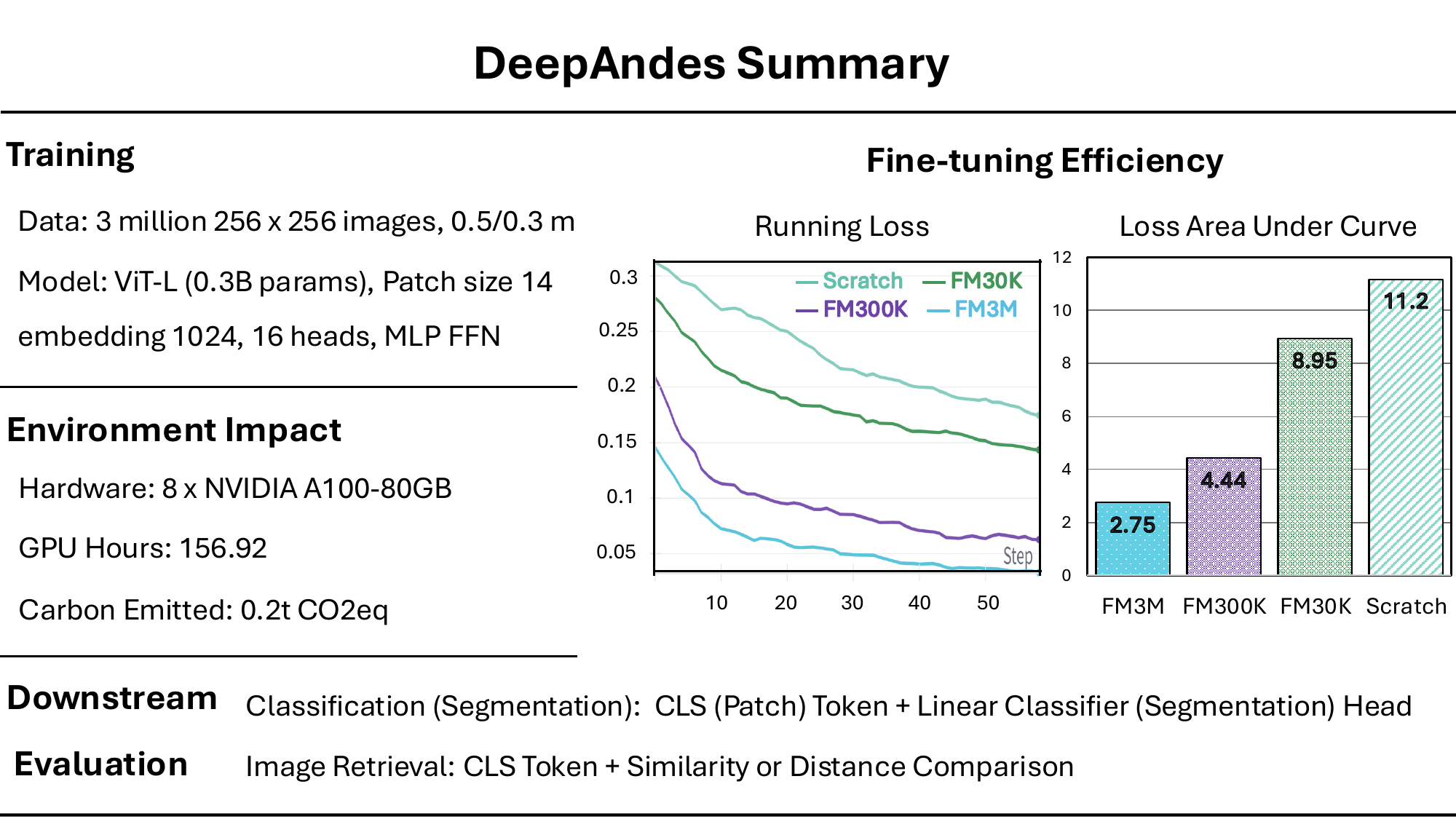}
\end{center}
\caption{\textbf{Summarization of DeepAndes Model and Fine-Tuning Efficiency Comparison}. The right panel illustrates the running loss experimental logs (10 epochs) from an imbalanced loci classification experiment, highlighting DeepAndes (FM3M)’s rapid convergence}
{ \label{fig:summary}}
\end{figure*}

Lastly, this section provides a summary of the pre-training and fine-tuning of DeepAndes (0.3B parameters vision transformer) in Figure~\ref{fig:summary}. Particularly, it includes training data, model, environmental impact, and evaluation strategies utilized in this work. To observe fine-tuning efficiency, we selected the imbalanced loci image classification task as representative. The first 10 epochs of the training log are displayed because the model tends to overfit beyond this point. As demonstrated, an increased pre-training scale promotes faster training convergence by monitoring the running loss and running loss AUC (the smaller, the better). This aligns with the foundation model's purpose to accommodate a broad range of tasks through few-shot or zero-shot learning.

\section{Discussion} 
The evaluation of downstream tasks indicates that DeepAndes enhances visual feature representations, which improve as pre-training scales increase. With the 3-million scale pre-training, DeepAndes achieves over 83\% accuracy in few-shot imbalance classification, surpassing other models, including those trained from scratch in a supervised manner, across various fine-tuning scales. Similar patterns are evident in image retrieval and segmentation tasks, where the feature embeddings from the pre-trained backbone show potential for both image-level and pixel-level remote sensing tasks when labeled data is limited. In addition to the main findings, it is important to consider the following aspects.

\subsection{Limitations}
\textcolor{black}{\textbf{Limited Pre-training and Data Curation.}}
A limitation of this work is that the data for self-supervised pre-training should undergo more thorough curation. In \cite{oquab2023dinov2}, the training data undergo a comprehensive data de-duplication process and employ image retrieval to refine the uncurated data source with curated sources such as ImageNet-22 \cite{imagenet}. In our work, we try to sample from diverse archaeological regions while scaling up the datasets, yet we lack curated datasets and detailed prior knowledge of the survey regions, unlike established benchmarks. The archaeological regions are not as distinct as these natural imaging datasets, suggesting a need for further de-duplication beyond excluding featureless samples. 

\textcolor{black}{\textbf{Inferior Experimental Outcomes.}}
Figure~\ref{fig:scale_law} (scaling law) shows that larger-scale pre-training yields greater benefits when labeled data are scarce. However, limitations remain. In \textbf{(1) imbalanced classification} (Table~\ref{tab:classification_performance}) with only 72 positive samples, performance gaps persist across pre-training scales (30K, 300K, 3M). In some contexts, obtaining millions of samples for pre-training can be difficult, and downstream tasks may contain even fewer positive samples, highlighting the need for improved pre-training and fine-tuning with imbalance-aware methods. For \textbf{(2) few-shot segmentation}, results show considerable noise (e.g., Figure~\ref{fig:seg1. seg2. seg3}, middle panel). Small foreground objects are particularly difficult, suggesting the object detection task may be more effective. Distinguishing active from archaeological corrals is also challenging when weathering obscures animal evidence, suggesting the integration of expert domain knowledge (e.g., elevation, coordinates, or textual data) to domain fine-tuning. For \textbf{(3) image-level content retrieval}, input queries with many loci types can yield noisy results, as features of archaeological interest may be overshadowed by others; pixel- or object-level retrieval will be needed in future work.

\textcolor{black}{\textbf{Feature Scaling Challenge.}} Similar to other Earth observation (EO) studies, features in our data vary greatly in scale, from 10-meter archaeological structures to corrals spanning hundreds of meters. As shown in the qualitative results of few-shot loci segmentation (Figure~\ref{fig:seg1. seg2. seg3}), using a 224-pixel input (the default for natural image benchmarks and many existing EO models) poses challenges: buildings appear as very small objects, while corrals may exceed the image crop size. In real archaeological surveys, researchers often switch between local and large-scale views to help them identify features of varying sizes and distributions. This suggests incorporating larger image crop sizes—i.e., higher pixel dimensions—during pre-training and specific fine-tuning. Larger image crop size can capture richer semantic information, potentially benefiting a wider range of downstream applications.

\subsection{Broad Impacts}
\textcolor{black}{\textbf{Value to Archaeology.}} Although costly, developing a foundation model can greatly improve our ability to identify archaeological features in high resolution satellite imagery, given the variability of landforms and archaeological features across the Andes. The few-shot results in classification and segmentation demonstrate a scaling law with increasing pre-training scale, confirming the effectiveness of the SSL framework. This framework provides a foundation model for diverse downstream tasks while reducing the burden of retraining and large-scale labeling. Additionally, through zero-shot image retrieval, the pre-trained backbone can rapidly mine and expand relevant datasets for specific archaeological objectives, a task infeasible at scale with manual methods.

\textcolor{black}{\textbf{Domain Generalization.}} While the proposed foundation model is explicitly pre-trained on extensive high-resolution satellite imagery from the Andes, its potential applications extend beyond this region. For archaeological patterns unique to the Andes, iterative fine-tuning with remote sensing imagery from additional areas could further enhance the model’s generalizability across a broader global span, supporting a wider range of computational applications. Beyond archaeology, the model may also advance geoscientific inquiries, including studies of land use, human–environment interaction, and cultural heritage preservation. In this sense, the Andes serve as a proving ground, demonstrating how large-scale pre-training can be well-suited for domain-specific applications such as archaeology. 

\subsection{Future Work}
For the future and in progress work, the DeepAndes will be integrated into the the GeoPACHA geospatial platform \cite{geopacha_paper} as a web application, collaborating regional experts and their teams. Meanwhile, depending on the remote sensing tasks, the human-in-the-loop is also included as experts will also verify and compare the autonomous model prediction with their data annotations. The curated data will be used for next round of foundation model fine-tuning to acquire more a specialized model, DeepAndesArch, for remote sensing tasks. From a model perspective, the proposed foundation model offers broad utility and can be further fine-tuned for supporting more advanced archaeological remote sensing tasks.

\section{Conclusion}
DeepAndes is the first AI vision foundation model for multi-spectral remote sensing data for social and earth science applications, utilizing the SOTA DINOv2 framework. DeepAndes is pre-trained on 3 million WorldView-2 and WorldView-3 satellite images with eight spectral bands. Through extensive downstream experiments on three prevalent computer vision tasks in archaeology—imbalanced image classification, image instance retrieval, and semantic segmentation. We demonstrates the effectiveness of DeepAndes for both image-level and pixel-level feature representation, as well as in few-shot learning capabilities. The pre-trained backbone will be integrated into the GeoPACHA web app to expand the scale of our archaeological surveys in the Andes with human-in-the-loop verification. The broad utility of the proposed foundation model can be further fine-tuned for more advanced archaeological remote sensing tasks. As we collect more data with the AI-assisted GeoPACHA tool, experts will be able to contribute more effectively as observers and analysts in the archaeological workflow.

\section*{Acknowledgements}
This work was supported by the following grants: National Endowment for the Humanities Level III Digital Enhancement Grant (Award ID Number HAA-293452-23), National Science Foundation IIS-III: Medium: Collaborative Research (Award number 2106717); National Science Foundation Collaborative Research: Research Infrastructure: HNDS-I (Award Numbers 2419793 and 2419794); Vanderbilt University Scaling Success Grant, Vanderbilt University Discovery Grant. The project also benefited from the infrastructural support of the Vanderbilt University Data Science Institute (GPU computational infrastructure) and the Vanderbilt University HLRB Lab (computational infrastructure), and the Spatial Analysis Research Laboratory (geospatial computational infrastructure). This manuscript has been co-authored by UT-Battelle LLC, under contract DE-AC05-00OR22725, with the Oak Ridge National Laboratory (ORNL) of the US Department of Energy (DOE). The U.S. Government retains a nonexclusive,worldwide license to publish or reproduce the published form of this manuscript, or to authorize others to do so, for U.S.Government purposes, as acknowledged by the publisher.


{\small
\bibliographystyle{ieee_fullname}
\bibliography{mybib}

\begin{thebibliography}{10}\itemsep=-1pt

\bibitem{akiva2022self}
Peri Akiva, Matthew Purri, and Matthew Leotta.
\newblock Self-supervised material and texture representation learning for remote sensing tasks.
\newblock In {\em Proceedings of the IEEE/CVF Conference on Computer Vision and Pattern Recognition}, pages 8203--8215, 2022.

\bibitem{proposal_8}
Susan Alcock and John Cherry.
\newblock {\em Side-by-Side Survey : Comparative Regional Studies in the Mediterranean World}.
\newblock Oxbow Books, 2016.

\bibitem{allwine2002overview}
K~Jerry Allwine, Joseph~H Shinn, Gerald~E Streit, Kirk~L Clawson, and Mike Brown.
\newblock Overview of urban 2000: A multiscale field study of dispersion through an urban environment.
\newblock {\em Bulletin of the American Meteorological Society}, 83(4):521--536, 2002.

\bibitem{awais2025foundation}
Muhammad Awais, Muzammal Naseer, Salman Khan, Rao~Muhammad Anwer, Hisham Cholakkal, Mubarak Shah, Ming-Hsuan Yang, and Fahad~Shahbaz Khan.
\newblock Foundation models defining a new era in vision: a survey and outlook.
\newblock {\em IEEE Transactions on Pattern Analysis and Machine Intelligence}, 2025.

\bibitem{ayush2021geography}
Kumar Ayush, Burak Uzkent, Chenlin Meng, Kumar Tanmay, Marshall Burke, David Lobell, and Stefano Ermon.
\newblock Geography-aware self-supervised learning.
\newblock In {\em Proceedings of the IEEE/CVF International Conference on Computer Vision}, pages 10181--10190, 2021.

\bibitem{proposal_9}
Andrew Bevan and James Conolly.
\newblock {\em Mediterranean islands, fragile communities and persistent landscapes: Antikythera in long-term perspective}.
\newblock Cambridge University Press, 2013.

\bibitem{proposal_10}
Brian~R. Billman and Gary~M. Feinman.
\newblock {\em Settlement pattern studies in the {Americas}: fifty years since {Virú}}.
\newblock Smithsonian series in archaeological inquiry. Smithsonian Institution press, Washington, D.C, 1999.

\bibitem{DINO}
Mathilde Caron, Hugo Touvron, Ishan Misra, Hervé Jégou, Julien Mairal, Piotr Bojanowski, and Armand Joulin.
\newblock Emerging properties in self-supervised vision transformers, 2021.

\bibitem{casana2014regional}
Jesse Casana.
\newblock Regional-scale archaeological remote sensing in the age of big data: Automated site discovery vs. brute force methods.
\newblock {\em Advances in Archaeological Practice}, 2(3):222--233, 2014.

\bibitem{SimCLR}
Ting Chen, Simon Kornblith, Mohammad Norouzi, and Geoffrey Hinton.
\newblock A simple framework for contrastive learning of visual representations, 2020.

\bibitem{SatMAE}
Yezhen Cong, Samar Khanna, Chenlin Meng, Patrick Liu, Erik Rozi, Yutong He, Marshall Burke, David Lobell, and Stefano Ermon.
\newblock Satmae: Pre-training transformers for temporal and multi-spectral satellite imagery.
\newblock In S. Koyejo, S. Mohamed, A. Agarwal, D. Belgrave, K. Cho, and A. Oh, editors, {\em Advances in Neural Information Processing Systems}, volume~35, pages 197--211. Curran Associates, Inc., 2022.

\bibitem{cui2024enhancing}
Can Cui, Ruining Deng, Junlin Guo, Quan Liu, Tianyuan Yao, Haichun Yang, and Yuankai Huo.
\newblock Enhancing physician flexibility: Prompt-guided multi-class pathological segmentation for diverse outcomes.
\newblock In {\em 2024 IEEE EMBS International Conference on Biomedical and Health Informatics (BHI)}, pages 1--8. IEEE, 2024.

\bibitem{cui2024pfps}
Can Cui, Ruining Deng, Junlin Guo, Quan Liu, Tianyuan Yao, Haichun Yang, and Yuankai Huo.
\newblock Pfps: Prompt-guided flexible pathological segmentation for diverse potential outcomes using large vision and language models.
\newblock {\em arXiv preprint arXiv:2407.09979}, 2024.

\bibitem{imagenet}
Jia Deng, Wei Dong, Richard Socher, Li-Jia Li, Kai Li, and Li Fei-Fei.
\newblock Imagenet: A large-scale hierarchical image database.
\newblock In {\em 2009 IEEE Conference on Computer Vision and Pattern Recognition}, pages 248--255, 2009.

\bibitem{Dias2023}
Philipe Dias, Abhishek Potnis, Sreelekha Guggilam, Lexie Yang, Aristeidis Tsaris, Henry Medeiros, and Dalton Lunga.
\newblock An agenda for multimodal foundation models for earth observation.
\newblock In {\em IGARSS 2023 - 2023 IEEE International Geoscience and Remote Sensing Symposium}, pages 1237--1240, 2023.

\bibitem{FAISS}
Matthijs Douze, Alexandr Guzhva, Chengqi Deng, Jeff Johnson, Gergely Szilvasy, Pierre-Emmanuel Mazar{\'e}, Maria Lomeli, Lucas Hosseini, and Herv{\'e} J{\'e}gou.
\newblock The faiss library.
\newblock {\em arXiv preprint arXiv:2401.08281}, 2024.

\bibitem{proposal_20}
Timothy~K Earle et~al.
\newblock Archaeological field research in the upper mantaro, peru, 1982-1983: Investigations of inka expansion and exchange.
\newblock {\em (No Title)}, 1987.

\bibitem{proposal_11}
Gary~M Feinman, Stephen~A Kowalewski, Laura Finsten, Richard~E Blanton, and Linda Nicholas.
\newblock Long-term demographic change: A perspective from the valley of oaxaca, mexico.
\newblock {\em Journal of Field Archaeology}, 12(3):333--362, 1985.

\bibitem{feng2023self}
Yingchao Feng, Peijin Wang, Wenhui Diao, Qibin He, Huiyang Hu, Hanbo Bi, Xian Sun, and Kun Fu.
\newblock A self-supervised cross-modal remote sensing foundation model with multi-domain representation and cross-domain fusion.
\newblock In {\em IGARSS 2023-2023 IEEE International Geoscience and Remote Sensing Symposium}, pages 2239--2242. IEEE, 2023.

\bibitem{fuller1998integration}
RM Fuller, GB Groom, S Mugisha, P Ipulet, D Pomeroy, A Katende, R Bailey, and R Ogutu-Ohwayo.
\newblock The integration of field survey and remote sensing for biodiversity assessment: a case study in the tropical forests and wetlands of sango bay, uganda.
\newblock {\em Biological conservation}, 86(3):379--391, 1998.

\bibitem{grizonnet2017orfeo}
Manuel Grizonnet, Julien Michel, Victor Poughon, Jordi Inglada, Micka{\"e}l Savinaud, and R{\'e}mi Cresson.
\newblock Orfeo toolbox: Open source processing of remote sensing images.
\newblock {\em Open Geospatial Data, Software and Standards}, 2(1):15, 2017.

\bibitem{guo2024good}
Junlin Guo, Siqi Lu, Can Cui, Ruining Deng, Tianyuan Yao, Zhewen Tao, Yizhe Lin, Marilyn Lionts, Quan Liu, Juming Xiong, et~al.
\newblock How good are we? evaluating cell ai foundation models in kidney pathology with human-in-the-loop enrichment.
\newblock {\em arXiv preprint arXiv:2411.00078}, 2024.

\bibitem{guo2025assessment}
Junlin Guo, Siqi Lu, Can Cui, Ruining Deng, Tianyuan Yao, Zhewen Tao, Yizhe Lin, Marilyn Lionts, Quan Liu, Juming Xiong, et~al.
\newblock Assessment of cell nuclei ai foundation models in kidney pathology.
\newblock In {\em Medical Imaging 2025: Image Perception, Observer Performance, and Technology Assessment}, volume 13409, pages 76--82. SPIE, 2025.

\bibitem{guo2024skysense}
Xin Guo, Jiangwei Lao, Bo Dang, Yingying Zhang, Lei Yu, Lixiang Ru, Liheng Zhong, Ziyuan Huang, Kang Wu, Dingxiang Hu, et~al.
\newblock Skysense: A multi-modal remote sensing foundation model towards universal interpretation for earth observation imagery.
\newblock In {\em Proceedings of the IEEE/CVF Conference on Computer Vision and Pattern Recognition}, pages 27672--27683, 2024.

\bibitem{MoCo}
Kaiming He, Haoqi Fan, Yuxin Wu, Saining Xie, and Ross Girshick.
\newblock Momentum contrast for unsupervised visual representation learning, 2020.

\bibitem{he2015deep}
Kaiming He, Xiangyu Zhang, Shaoqing Ren, and Jian Sun.
\newblock Deep residual learning for image recognition, 2015.

\bibitem{hong2024spectralgpt}
Danfeng Hong, Bing Zhang, Xuyang Li, Yuxuan Li, Chenyu Li, Jing Yao, Naoto Yokoya, Hao Li, Pedram Ghamisi, Xiuping Jia, et~al.
\newblock Spectralgpt: Spectral remote sensing foundation model.
\newblock {\em IEEE Transactions on Pattern Analysis \& Machine Intelligence}, 46(08):5227--5244, 2024.

\bibitem{Jiao2023}
Licheng Jiao, Zhongjian Huang, Xiaoqiang Lu, Xu Liu, Yuting Yang, Jiaxuan Zhao, Jinyue Zhang, Biao Hou, Shuyuan Yang, Fang Liu, Wenping Ma, Lingling Li, Xiangrong Zhang, Puhua Chen, Zhixi Feng, Xu Tang, Yuwei Guo, Dou Quan, Shuang Wang, Weibin Li, Jing Bai, Yangyang Li, Ronghua Shang, and Jie Feng.
\newblock Brain-inspired remote sensing foundation models and open problems: A comprehensive survey.
\newblock {\em IEEE Journal of Selected Topics in Applied Earth Observations and Remote Sensing}, 16:10084--10120, 2023.

\bibitem{jing2019selfsupervised}
Longlong Jing and Yingli Tian.
\newblock Self-supervised visual feature learning with deep neural networks: A survey, 2019.

\bibitem{ke2024tshfna}
Jing Ke, Junchao Zhu, Xin Yang, Haolin Zhang, Yuxiang Sun, Jiayi Wang, Yizhou Lu, Yiqing Shen, Sheng Liu, Fusong Jiang, et~al.
\newblock Tshfna-examiner: A nuclei segmentation and cancer assessment framework for thyroid cytology image.
\newblock {\em Journal of Shanghai Jiaotong University (Science)}, 29(6):945--957, 2024.

\bibitem{leevy2018survey}
Joffrey~L Leevy, Taghi~M Khoshgoftaar, Richard~A Bauder, and Naeem Seliya.
\newblock A survey on addressing high-class imbalance in big data.
\newblock {\em Journal of Big Data}, 5(1):1--30, 2018.

\bibitem{li2024u}
Chenxin Li, Xinyu Liu, Wuyang Li, Cheng Wang, Hengyu Liu, Yifan Liu, Zhen Chen, and Yixuan Yuan.
\newblock U-kan makes strong backbone for medical image segmentation and generation.
\newblock {\em arXiv preprint arXiv:2406.02918}, 2024.

\bibitem{li2025unleashing}
Yansheng Li, Jieyi Tan, Bo Dang, Mang Ye, Sergey~A Bartalev, Stanislav Shinkarenko, Linlin Wang, Yingying Zhang, Lixiang Ru, Xin Guo, et~al.
\newblock Unleashing the potential of remote sensing foundation models via bridging data and computility islands.
\newblock {\em The Innovation}, 2025.

\bibitem{liu2024remoteclip}
Fan Liu, Delong Chen, Zhangqingyun Guan, Xiaocong Zhou, Jiale Zhu, Qiaolin Ye, Liyong Fu, and Jun Zhou.
\newblock Remoteclip: A vision language foundation model for remote sensing.
\newblock {\em IEEE Transactions on Geoscience and Remote Sensing}, 2024.

\bibitem{lu2025vision}
Siqi Lu, Junlin Guo, James~R Zimmer-Dauphinee, Jordan~M Nieusma, Xiao Wang, Steven~A Wernke, Yuankai Huo, et~al.
\newblock Vision foundation models in remote sensing: A survey.
\newblock {\em IEEE Geoscience and Remote Sensing Magazine}, 2025.

\bibitem{Ma2024}
Yuchi Ma, Shuo Chen, Stefano Ermon, and David~B. Lobell.
\newblock Transfer learning in environmental remote sensing.
\newblock {\em Remote Sensing of Environment}, 301:113924, 2024.

\bibitem{SatMAE++}
Mubashir Noman, Muzammal Naseer, Hisham Cholakkal, Rao~Muhammad Anwar, Salman Khan, and Fahad~Shahbaz Khan.
\newblock Rethinking transformers pre-training for multi-spectral satellite imagery, 2024.

\bibitem{oquab2023dinov2}
Maxime Oquab, Timoth{\'e}e Darcet, Th{\'e}o Moutakanni, Huy Vo, Marc Szafraniec, Vasil Khalidov, Pierre Fernandez, Daniel Haziza, Francisco Massa, Alaaeldin El-Nouby, et~al.
\newblock Dinov2: Learning robust visual features without supervision.
\newblock {\em arXiv preprint arXiv:2304.07193}, 2023.

\bibitem{proposal_21}
Jeffrey~R Parsons, Charles~M Hastings, and Ramiro Matos.
\newblock {\em Prehispanic Settlement Patterns in the Upper Mantaro and Tarma Drainages, Jun{\'\i}n, Peru: The Tarama-Chinchaycocha Region, Vol. 1, Parts 1 and 2}, volume~34.
\newblock U OF M MUSEUM ANTHRO ARCHAEOLOGY, 2000.

\bibitem{Scale-MAE}
Colorado~J. Reed, Ritwik Gupta, Shufan Li, Sarah Brockman, Christopher Funk, Brian Clipp, Kurt Keutzer, Salvatore Candido, Matt Uyttendaele, and Trevor Darrell.
\newblock Scale-mae: A scale-aware masked autoencoder for multiscale geospatial representation learning, 2023.

\bibitem{sakai2024ai_pnas}
Masato Sakai, Akihisa Sakurai, Siyuan Lu, Jorge Olano, Conrad~M Albrecht, Hendrik~F Hamann, and Marcus Freitag.
\newblock Ai-accelerated nazca survey nearly doubles the number of known figurative geoglyphs and sheds light on their purpose.
\newblock {\em Proceedings of the National Academy of Sciences}, 121(40):e2407652121, 2024.

\bibitem{sharma2021machine}
Neha Sharma, Reecha Sharma, and Neeru Jindal.
\newblock Machine learning and deep learning applications-a vision.
\newblock {\em Global Transitions Proceedings}, 2(1):24--28, 2021.

\bibitem{shi2023long}
Jiang-Xin Shi, Tong Wei, Zhi Zhou, Jie-Jing Shao, Xin-Yan Han, and Yu-Feng Li.
\newblock Long-tail learning with foundation model: Heavy fine-tuning hurts.
\newblock {\em arXiv preprint arXiv:2309.10019}, 2023.

\bibitem{tripcevich2010site}
Nicholas Tripcevich and Steven~A Wernke.
\newblock On-site recording of excavation data using mobile gis.
\newblock {\em Journal of Field Archaeology}, 35(4):380--397, 2010.

\bibitem{vanvalkenburgh2020big}
Parker VanValkenburgh and J~Andrew Dufton.
\newblock Big archaeology: Horizons and blindspots, 2020.

\bibitem{vaswani2023attention}
Ashish Vaswani, Noam Shazeer, Niki Parmar, Jakob Uszkoreit, Llion Jones, Aidan~N. Gomez, Lukasz Kaiser, and Illia Polosukhin.
\newblock Attention is all you need, 2023.

\bibitem{vorontsov2023virchow}
Eugene Vorontsov, Alican Bozkurt, Adam Casson, George Shaikovski, Michal Zelechowski, Siqi Liu, Kristen Severson, Eric Zimmermann, James Hall, Neil Tenenholtz, et~al.
\newblock Virchow: A million-slide digital pathology foundation model.
\newblock {\em arXiv preprint arXiv:2309.07778}, 2023.

\bibitem{DINO-MC}
Xinye Wanyan, Sachith Seneviratne, Shuchang Shen, and Michael Kirley.
\newblock Extending global-local view alignment for self-supervised learning with remote sensing imagery, 2024.

\bibitem{geopacha_paper}
Steven~A Wernke, Parker Van~Valkenburgh, James Zimmer-Dauphinee, Bethany Whitlock, Giles~Spence Morrow, Ryan Smith, Douglas Smit, Grecia~Roque Ortega, Kevin~Ricci Jara, Daniel Plekhov, et~al.
\newblock Large-scale, collaborative imagery survey in archaeology: the geospatial platform for andean culture, history and archaeology (geopacha).
\newblock {\em Antiquity}, 98(397):155--171, 2024.

\bibitem{proposal_12}
Gordon~R Willey.
\newblock Prehistoric settlement patterns in the vir{\'u}; valley, peru.
\newblock {\em Bureau of American Ethnology Bulletin}, 1953.

\bibitem{proposal_19}
David~John Wilson.
\newblock Prehispanic settlement patterns in the lower santa valley, peru: a regional perspective on the origins and development of complex north coast society.
\newblock {\em (No Title)}, 1988.

\bibitem{ssl_prior}
Jiachen Xu, Junlin Guo, James Zimmer-Dauphinee, Quan Liu, Yuxuan Shi, Zuhayr Asad, D~Mitchell Wilkes, Parker VanValkenburgh, Steven~A Wernke, and Yuankai Huo.
\newblock Semi-supervised contrastive learning for remote sensing: identifying ancient urbanization in the south-central andes.
\newblock {\em International journal of remote sensing}, 44(6):1922--1938, 2023.

\bibitem{yue2025glofinder}
Jialin Yue, Tianyuan Yao, Ruining Deng, Siqi Lu, Junlin Guo, Quan Liu, Juming Xiong, Mengmeng Yin, Haichun Yang, and Yuankai Huo.
\newblock Glofinder: Ai-empowered qupath plugin for wsi-level glomerular detection, visualization, and curation.
\newblock {\em Journal of Pathology Informatics}, page 100433, 2025.

\bibitem{zhou2024comprehensive}
Ce Zhou, Qian Li, Chen Li, Jun Yu, Yixin Liu, Guangjing Wang, Kai Zhang, Cheng Ji, Qiben Yan, Lifang He, et~al.
\newblock A comprehensive survey on pretrained foundation models: A history from bert to chatgpt.
\newblock {\em International Journal of Machine Learning and Cybernetics}, pages 1--65, 2024.

\bibitem{zhou2021ibot}
Jinghao Zhou, Chen Wei, Huiyu Wang, Wei Shen, Cihang Xie, Alan Yuille, and Tao Kong.
\newblock ibot: Image bert pre-training with online tokenizer.
\newblock {\em arXiv preprint arXiv:2111.07832}, 2021.

\bibitem{zhu2024asign}
Junchao Zhu, Ruining Deng, Tianyuan Yao, Juming Xiong, Chongyu Qu, Junlin Guo, Siqi Lu, Mengmeng Yin, Yu Wang, Shilin Zhao, et~al.
\newblock Asign: An anatomy-aware spatial imputation graphic network for 3d spatial transcriptomics.
\newblock {\em arXiv preprint arXiv:2412.03026}, 2024.

\end{thebibliography}
}
 
\newpage
\section{Biography Section}
\vspace{-30pt}
\begin{IEEEbiographynophoto}{Junlin Guo}
(junlin.guo@vanderbilt.edu) is a PhD student with the Biomedical Data Representation and Learning Lab (HRLB) and the Spatial Analysis Research Laboratory (SARL) at Vanderbilt University. His research interests include deep learning, foundation models, and their applications in remote sensing and medical image analysis.
\end{IEEEbiographynophoto}

\vspace{-30pt}
\begin{IEEEbiographynophoto}{James R Zimmer-Dauphinee} 
(james.r.zimmer-dauphinee@vanderbilt\\.edu) 
is a postdoctoral researcher with the Spatial Analysis Research Laboratory (SARL) at Vanderbilt University. His research interests include developing deep learning models for large-scale autonomous archaeological satellite imagery surveys, geophysical methods, and spatial modeling to understand the impact of colonization on indigenous peoples.
\end{IEEEbiographynophoto}

\vspace{-30pt}
\begin{IEEEbiographynophoto}{Jordan M Nieusma}
(jordan.m.nieusma@vanderbilt.edu)
is a data scientist and research assistant with the Spatial Analysis Research Laboratory (SARL) at Vanderbilt University. Her research interests include data science, deep learning and efficient learning of large foundation models. 
\end{IEEEbiographynophoto}

\vspace{-30pt}
\begin{IEEEbiographynophoto}{Siqi Lu}
(siqi.lu@vanderbilt.edu) is a Master's student with the Biomedical Data Representation and Learning Lab (HRLB) and the Spatial Analysis Research Laboratory (SARL) at Vanderbilt University. Her research interests include deep learning, medical image analysis, and software engineering.  
\end{IEEEbiographynophoto}

\vspace{-30pt}
\begin{IEEEbiographynophoto}{Quan Liu}
(quan.liu@vanderbilt.edu) is a researcher with the Biomedical Data Representation and Learning Lab (HRLB) at Vanderbilt University. His research interests include deep learning, artificial intelligence, and medical image analysis. 
\end{IEEEbiographynophoto}

\vspace{-30pt}
\begin{IEEEbiographynophoto}{Ruining Deng}
(rud4004@med.cornell.edu) is a research fellow with Weill Cornell Medicine in New York and the Biomedical Data Representation and Learning Lab (HRLB) at Vanderbilt University. His research interests include deep learning, artificial intelligence, and medical image analysis, which aim to explore clinical knowledge and assist diagnosis in a data-driven way.
\end{IEEEbiographynophoto}

\vspace{-30pt}
\begin{IEEEbiographynophoto}{Can Cui}
(can.cui.1@vanderbilt.edu) is a researcher with the Biomedical Data Representation and Learning Lab (HRLB) at Vanderbilt University. Her research interests include deep learning, data science, and medical image analysis, which aims for multi-modal learning for disease diagnosis and prognosis.
\end{IEEEbiographynophoto}

\vspace{-30pt}
\begin{IEEEbiographynophoto}{Jialin Yue}
(jialin.yue@vanderbilt.edu) is a Master's student with the Biomedical Data Representation and Learning Lab (HRLB) at Vanderbilt University. Her research interests include medical image processing and machine learning.
\end{IEEEbiographynophoto}

\vspace{-30pt}
\begin{IEEEbiographynophoto}{Yizhe Lin}
(yizhe.lin@vanderbilt.edu) is a undergraduate research assistant with the Spatial Analysis Research Laboratory (SARL) at Vanderbilt University.
\end{IEEEbiographynophoto}

\vspace{-30pt}
\begin{IEEEbiographynophoto}{Tianyuan Yao}
(tianyuan.yao@vanderbilt.edu) is a PhD student with the Biomedical Data Representation and Learning Lab (HRLB) at Vanderbilt University. His research interests include medical image analysis, deep learning, computer vision and their applications in pathology and radiology imaging.
\end{IEEEbiographynophoto}

\vspace{-30pt}
\begin{IEEEbiographynophoto}{Juming Xiong}
(juming.xiong@vanderbilt.edu) is a PhD student with the Biomedical Data Representation and Learning Lab (HRLB) at Vanderbilt University. His research interests include medical image analysis, deep learning, and data science.
\end{IEEEbiographynophoto}

\vspace{-30pt}
\begin{IEEEbiographynophoto}{Junchao Zhu}
(junchao.zhu@vanderbilt.edu) is a PhD student with the Biomedical Data Representation and Learning Lab (HRLB) at Vanderbilt University. His research interests include medical image analysis, deep learning, computer vision and their applications in large-scale pathology image processing.
\end{IEEEbiographynophoto}

\vspace{-30pt}
\begin{IEEEbiographynophoto}{Chongyu Qu}
(chongyu.qu@vanderbilt.edu) is a PhD student with the Biomedical Data Representation and Learning Lab (HRLB) at Vanderbilt University. His research interests include medical image analysis, deep learning, computer vision and post-training model quantization for large scale medical vision tasks.
\end{IEEEbiographynophoto}

\vspace{-30pt}
\begin{IEEEbiographynophoto}{Yuechen Yang}
(yuechen.yang@vanderbilt.edu) is a PhD student with the Biomedical Data Representation and Learning Lab (HRLB) at Vanderbilt University. Her research interests include medical image analysis, data science, and medical image processing, with a particular focus on Pathomics.
\end{IEEEbiographynophoto}

\vspace{-30pt}
\begin{IEEEbiographynophoto}{Mitchell Wilkes}
(mitch.wilkes@vanderbilt.edu) is an Associate Professor of Electrical and Computer Engineering at Vanderbilt University. Dr. Wilkes's research focuses on digital signal processing, image processing and computer vision, digital signal processing hardware, structurally adaptive systems, sonar, and signal modeling. Dr. Wilkes's intellectual neighborhoods also include Biomedical Imaging and Biophotonics, Surgery, and Engineering.
\end{IEEEbiographynophoto}

\vspace{-30pt}
\begin{IEEEbiographynophoto}{Xiao Wang}
(wangx2@ornl.gov) is a research staff scientist at Oak Ridge National Laboratory. His research interests include applying machine learning, medical physics, image processing, and high-performance computing to various imaging problems. He was awarded the HPCWire Supercomputing Achievement Award in 2024.
\end{IEEEbiographynophoto}

\vspace{-30pt}
\begin{IEEEbiographynophoto}{Parker VanValkenburgh}
(parker\_vanvalkenburgh@brown.edu) is an Associate Professor of Anthropology and Archaeology at Brown University. Dr. VanValkenburgh's research focuses on the impacts of colonialism and imperialism on Indigenous people and environments in the Peruvian Andes. He utilizes diverse materials and digital methodologies, including GIS, to understand the transformation of relationships in imperial histories. Dr.VanValkenburgh co-directs the Paisajes Arqueológicos de Chachapoyas (PACha) project and GeoPACHA (Geospatial Platform for Andean Culture, History, and Archaeology).
\end{IEEEbiographynophoto}

\vspace{-30pt}
\begin{IEEEbiographynophoto}{Steven A Wernke}
(s.wernke@vanderbilt.edu)
is Professor and Chair of Anthropology at Vanderbilt University, director of the Spatial Analysis Research Laboratory (SARL), and director of the Vanderbilt Institute for Spatial Research. Dr. Wernke is an archaeologist and historical anthropologist of the Andean region of South America. His research combines archaeology and history, prehispanic and colonial studies, as well as anthropology and cultural geography. His interests center on the lived experiences of Indigenous communities across the Spanish invasion of the Americas, and on long-term, large-scale networks, social formations, and human-environment interactions across the Andes.
\end{IEEEbiographynophoto}

\vspace{-30pt}
\begin{IEEEbiographynophoto}{Yuankai Huo}
(yuankai.huo@vanderbilt.edu) is an Assistant Professor of Computer Science, and Electrical and Computer Engineering, as well as the Director of the Biomedical Data Representation and Learning Lab (HRLB Lab) at Vanderbilt University. Additionally, he is an Assistant Professor of Pathology, Microbiology, and Immunology at Vanderbilt University Medical Center. Dr. Huo’s current research specializes in high-dimensional multi-modal data analysis, computational pathology and radiology, and medical computer vision. Dr. Huo has received prestigious awards, including the Charles E. Ives Journal Award from the Society for Imaging Science and Technology, the Early Career Achievement Award from the Society for Imaging Informatics in Medicine, and the NAIRR Pilot award from NSF. He is a Senior Member of IEEE and a lifetime member of SPIE, contributing as an organization committee member and area chair for leading medical image analysis conferences such as MICCAI, MIDL, and ISBI. His ongoing efforts are dedicated to advancing next-generation AI algorithms for ultra-high-resolution imaging and non-imaging data analysis.
\end{IEEEbiographynophoto}

\vfill

\end{document}